%% file: main.tex
\documentclass{article}

\usepackage{microtype}
\usepackage{graphicx}
\usepackage{subcaption}
\usepackage{placeins}
\usepackage{booktabs} 
\usepackage{tcolorbox}
\usepackage{listings}  

\newtcolorbox{promptbox}[1][]{
    colback=gray!5,
    colframe=gray!60,
    fonttitle=\bfseries,
    title=#1,
    boxrule=0.5pt,
    arc=2pt,
    left=6pt,
    right=6pt,
    top=6pt,
    bottom=6pt
}

\usepackage[hyphens]{url}
\usepackage{hyperref}

\usepackage[preprint]{icml2026}

\usepackage{amsmath}
\usepackage{amssymb}
\usepackage{mathtools}
\usepackage{amsthm}

\usepackage[capitalize,noabbrev]{cleveref}

\theoremstyle{plain}

\theoremstyle{definition}

\theoremstyle{remark}

\usepackage{enumitem}
\setlist{itemsep=2pt, topsep=2pt, parsep=0pt, partopsep=0pt}

\usepackage[textsize=tiny]{todonotes}

\icmltitlerunning{Efficiently Aligning Language Models with Online Natural Language Feedback}

\begin{document}

\twocolumn[
  \icmltitle{Efficiently Aligning Language Models with Online Natural Language Feedback}
  \icmlsetsymbol{equal}{*}

  \begin{icmlauthorlist}
    \icmlauthor{Christine Ye}{stanford,afp}
    \icmlauthor{Joe Benton}{anthropic}
  \end{icmlauthorlist}

  \icmlaffiliation{stanford}{Stanford University}
  \icmlaffiliation{afp}{Anthropic Fellows Program for Safety Research}
  \icmlaffiliation{anthropic}{Anthropic}

  \icmlcorrespondingauthor{Christine Ye}{cye@stanford.edu}
  \icmlcorrespondingauthor{Joe Benton}{joe@anthropic.com}

  \icmlkeywords{Machine Learning, ICML}

  \vskip 0.3in
]



\printAffiliationsAndNotice{}  

\begin{abstract}
\input{Sections/0_abstract}
\end{abstract}

\section{Introduction}
\input{Sections/1_introduction}

\section{Preliminaries and Setup}
\input{Sections/2_preliminaries}

\section{Online feedback via in-context learning} \label{sec:icl}
\input{Sections/3_icl}

\section{Online feedback via fine-tuning}\label{sec:ft}
\input{Sections/4_finetuning}

\section{Understanding usefulness of proxy rewards}\label{sec:alignment}
\input{Sections/5_alignment}

\section{Related Work}
\input{Sections/6_related_work}

\section{Discussion}
\input{Sections/7_discussion}

\section*{Acknowledgments}
CY would like to thank Alex Cloud, Hailey Joren, Neil Rathi, and Claire Short for their very helpful input.

\bibliographystyle{icml2026}  
\bibliography{example_paper}  

\newpage
~
\newpage

\onecolumn

\appendix
\input{Sections/8_appendix}

\section{Prompts}
\label{sec:prompts}
\input{Sections/9_prompts}

\end{document}

%% file: Sections/0_abstract.tex
Reinforcement learning with verifiable rewards has been used to elicit impressive performance from language models in many domains. But, broadly beneficial deployments of AI may require us to train models with strong capabilities in ``fuzzy'', hard-to-supervise domains. In this paper, we develop methods to align language models in fuzzy domains where human experts are still able to provide high-quality supervision signal, but only for a small number of model outputs, using \textit{online natural language feedback}. Specifically, we train models by iteratively optimizing against proxy reward signals, stopping at the point of over-optimization, collecting fresh expert supervision, and updating the proxy reward. We construct proxy reward models from language models using in-context learning (ICL) and fine-tuning. We test our methods by eliciting creative writing and alignment research capabilities in Qwen3-8B and Haiku 4.5 respectively. For Qwen3-8B, ICL methods recover up to 35\% of performance with 50x fewer expert samples, while fine-tuning methods recover 80\% with up to 20x fewer samples and 100\% with 3x fewer samples. For Haiku 4.5, ICL methods recover up to 35\% of performance with 30x fewer samples, and fine-tuning methods recover 100\% with 10x fewer samples. Our results suggest that online natural language feedback can substantially improve the data efficiency of expert supervision.

%% file: Sections/1_introduction.tex
Training large language models (LLMs) using reinforcement learning with verifiable rewards (RLVR) has substantially improved their performance in domains such as math reasoning, logic puzzles, formalization, and code generation \citep{Guo_2025, wen2025reinforcementlearningverifiablerewards, xin2024deepseekproverv15harnessingproofassistant}. RLVR uses a verifier (for example, unit test cases or ground truth problem answers) to grade the correctness of the model's outputs, allowing cheap and high-quality reward specification. When verifiers are available, scaling supervision is relatively straightforward. 

Beyond verifiable tasks, though, it is important that we can elicit language models' capabilities on ``fuzzy'' tasks, or tasks which cannot be easily and objectively verified. Many beneficial applications of powerful AI models, such as alignment research, diplomacy, or moral philosophy, disproportionately require models to have strong capabilities in difficult-to-supervise domains. But unlike in verifiable settings, it remains unclear how we can scale up to and beyond human expert supervision for ``fuzzy'' tasks. This is the \textbf{scalable oversight} problem, or the question of how to supervise models whose capabilities may surpass those of humans \citep{leike2018scalableagentalignmentreward, bowman2022measuringprogressscalableoversight}.

Much existing work on scalable oversight focuses on the case where supervising model outputs is too \textit{difficult} for any human, motivating methods like debate and consultancy \citep{khan2024debatingpersuasivellmsleads, kenton2024scalableoversightweakllms}. We focus on the regime where humans could still feasibly supervise models, but where models generate outputs at speeds and scales that make human supervision intractable. Consider the case where only a small group of humans can provide sufficiently high-quality supervision, or where full evaluation of each output requires a prohibitive amount of time \citep{shlegeris2024scalable}. Training models with this supervision would require substantial improvements in data efficiency. 

In this work, we develop data-efficient training protocols that iteratively use natural language expert feedback to update a proxy reward model, and use the proxy reward model to supervise training. To correct over-optimization against the proxy during training \citep{gao2022scalinglawsrewardmodel, wen2024languagemodelslearnmislead}, we implement iterated online RLHF \citep{christiano2017deep, bai2022traininghelpfulharmlessassistant, guo2024directlanguagemodelalignment}:
\begin{enumerate}
    \item RL against a proxy reward until we over-optimize with respect to the ground truth reward.
    \item Stop and collect natural language expert feedback.
    \item Efficiently update the proxy reward model by distilling the natural language feedback.
\end{enumerate}
As previous work has shown that natural language feedback provides richer supervision than a preference pair or scalar reward alone, improving data efficiency for some use cases \citep{scheurer2022training, yang2024largelanguagemodelsoptimizers, agrawal2025gepareflectivepromptevolution, lee2025feedbackdescentopenendedtext}, we use the natural language feedback to efficiently update the proxy rewards.

Following a sandwiching approach \citep{cotra2021case, bowman2022measuringprogressscalableoversight}, we conduct experiments in a scaled-down setting analogous to humans supervising stronger models:
\begin{itemize}
    \item An \textbf{expert} provides high-quality supervision through long-form feedback and scalar rewards on individual rollouts. We substitute the human expert with a strong frontier LLM.
    \item The \textbf{policy} is a substantially smaller and weaker model than the expert, analogous to the capability gap between human experts and the models they train.
    \item A \textbf{proxy reward model} of similar strength to the policy assigns reward during RL; we base the proxy reward model off the policy model.
\end{itemize}

Following \citet{burns2023weaktostrong}, we measure \textit{effectiveness} as the \textit{performance gap recovered} (PGR), the fraction of the gap between maximum elicitable performance and baseline that our methods recover. We measure \textit{data efficiency} as the number of expert samples used.

\begin{figure*}[t]
    \centering
    \includegraphics[width=0.65\textwidth]{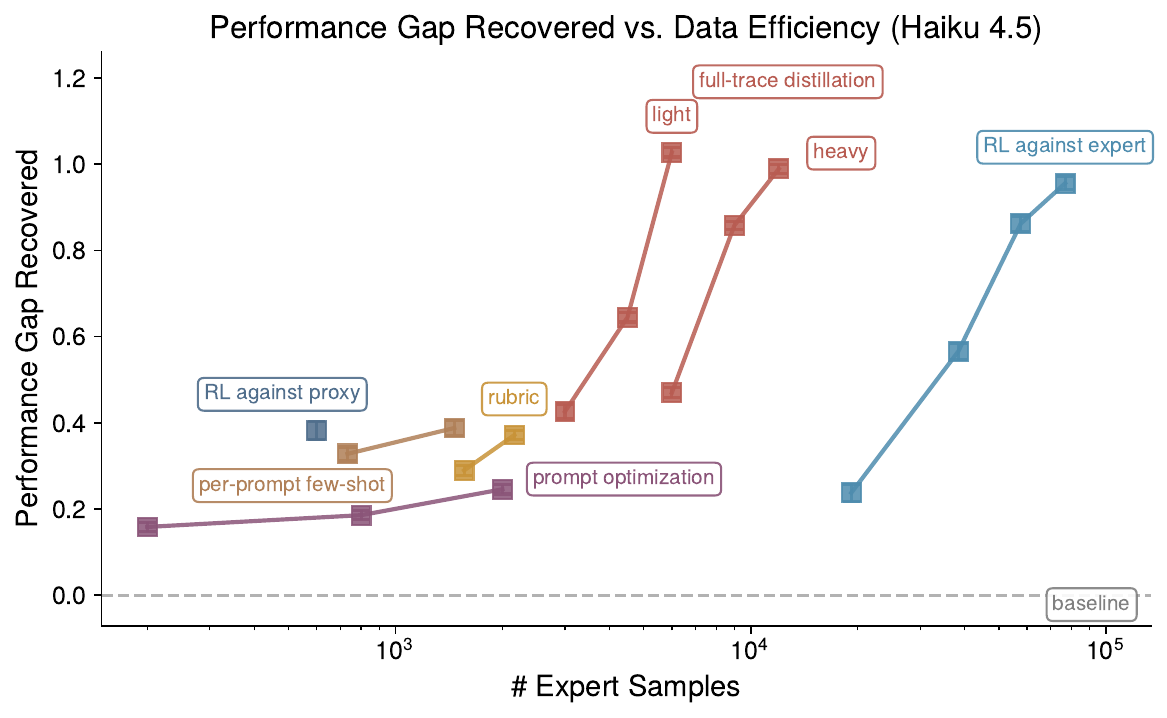}
    \caption{(Haiku 4.5 setting) Performance gap recovered and number of expert samples required by our online natural language feedback training methods, tested on eliciting Haiku 4.5 to write alignment research experiment plans.}
    \label{fig:fig1_haiku}
\end{figure*}

\begin{figure*}[t]
    \centering
    \includegraphics[width=0.65\textwidth]{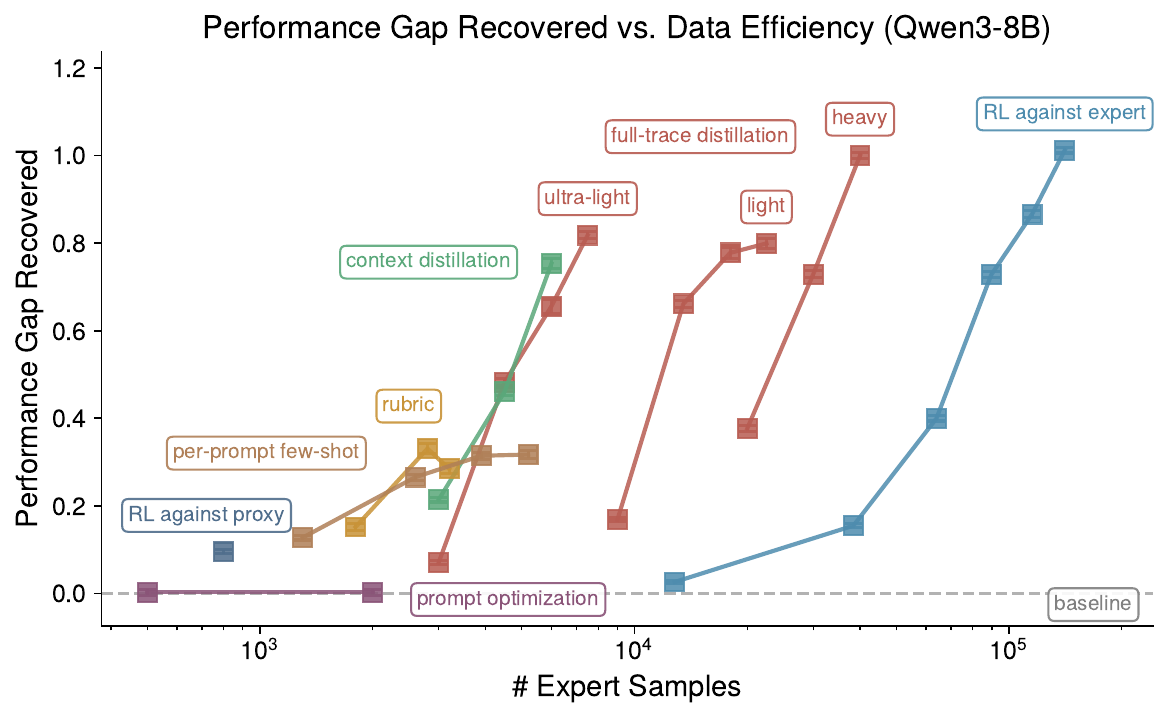}
    \caption{(Qwen3-8B setting) Performance gap recovered and number of expert samples required by our online natural language feedback training methods, tested on eliciting Qwen3-8B to write short stories.}
    \label{fig:fig1}
\end{figure*}

We study protocols which update the proxy reward model via in-context learning (Section \ref{sec:icl}) or fine-tuning (Section \ref{sec:ft}). We also discuss considerations for measuring or predicting the robustness of proxy reward models to over-optimization (Section \ref{sec:alignment}). Our findings, shown in Figures \ref{fig:fig1_haiku} and \ref{fig:fig1}, are:
\begin{itemize}
    \item Protocols relying on in-context learning are efficient but not fully effective. For Qwen3-8B, we recover up to 35\% of performance with 50x fewer expert samples. For Haiku 4.5, we recover up to 35\% of performance using 30x fewer expert samples, although iterative protocols do not out-perform our baselines. We spend most of our sample budget on tracking over-optimization and selecting between proxy rewards.
    \item Protocols using supervised fine-tuning are highly effective and somewhat efficient. For Qwen3-8B, we recover around 80\% of performance while requiring up to 20x fewer expert samples, and 100\% of performance using 3x fewer samples. For Haiku 4.5, we fully recover performance using 10x fewer expert samples. Full natural language feedback is important: proxy reward models trained only on the expert's scalar rewards are much less robust.
    \item The static alignment between proxy and expert rewards, as measured by advantage correlation, can be a useful heuristic for interpreting performance.
\end{itemize}

%% file: Sections/2_preliminaries.tex
We work in two ``sandwiched'' settings, distinct in task and model scale: training Qwen3-8B to write creative short stories with Claude Opus 4.1 expert supervision, and training Claude Haiku 4.5 to write alignment research experiment plans with Claude Opus 4.5 expert supervision.

\subsection{Creative writing setting (Qwen3-8B)}

\textbf{Data}\hspace{3mm} The task is to write 1000-word creative short stories, a highly subjective and decidedly difficult-to-verify task. We use a training set of 500 prompts templated with a 1-5 word story topic suggestion, and an evaluation set of 150 prompts from the same distribution. To collect expert feedback, we use a simple, open-ended prompt tailored to creative writing, which prompts the expert model to produce detailed reasoning about the story's merits, then deliberate and produce a scalar score. We count all this reasoning as the natural language feedback. (See Appendix \ref{sec:prompts} for details.)

\textbf{Models}\hspace{3mm} We use Qwen3-8B \citep{yang2025qwen3} in thinking mode as our policy and proxy reward model, and Claude Opus 4.1 \cite{anthropic2025opus41} without extended thinking as our expert model. On a single grading task, the scalar rewards from independent grading traces from Opus 4.1 are somewhat noisy, with standard deviation of 4-5 points (out of 100). 

\subsection{Alignment research setting (Haiku 4.5)}

\textbf{Data}\hspace{3mm} The  task is to write a detailed, implementable 1000-word plan describing the best initial experiment for a given alignment research proposal. We collect a corpus of empirical alignment research project proposals from previous batches of ML Alignment and Theory Scholars and the Anthropic Fellows Program for Safety Research, for a total of approximately 300 train and 60 test proposals. To collect expert feedback, we use a simple, open-ended prompt tailored to alignment research, which prompts the model to reason, produce a detailed report, and deliberate before assigning a score. (See Appendix \ref{sec:prompts} for details.)

\begin{figure}[tb]
    \centering
    \begin{subfigure}[t]{\linewidth}
        \centering
        \includegraphics[width=0.8\linewidth]{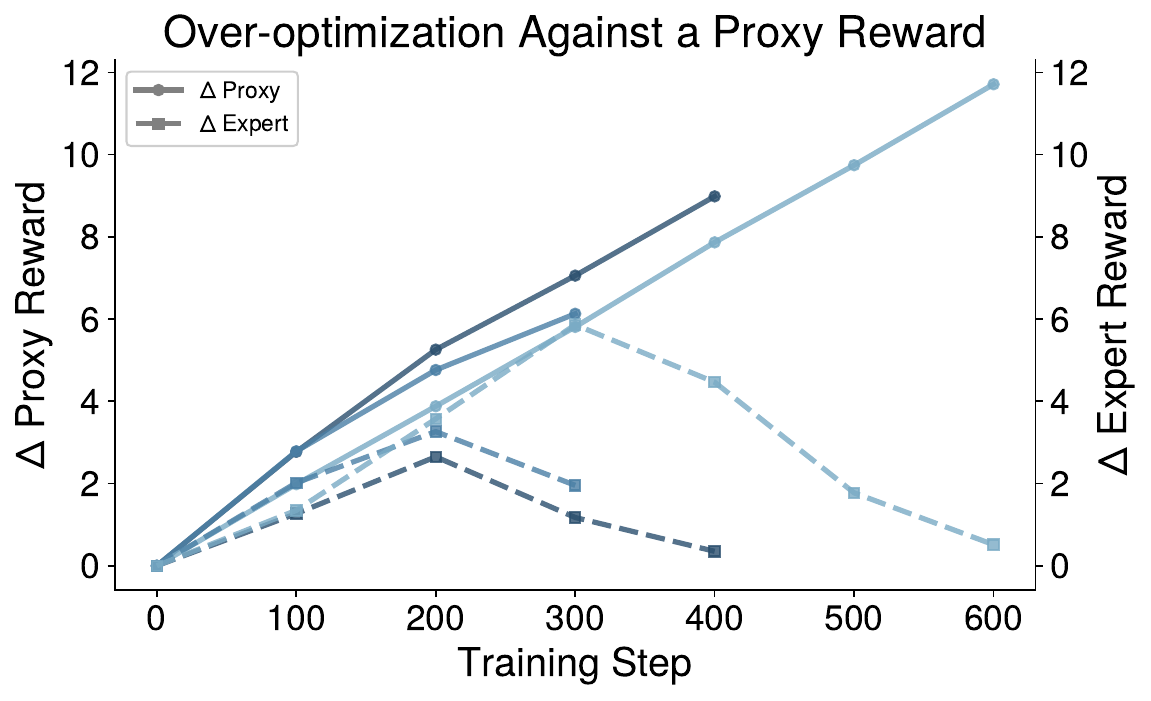}
        \caption{Qwen3-8B setting.}
        \label{fig:baselines}
    \end{subfigure}
    \\[1ex]
    \begin{subfigure}[t]{\linewidth}
        \centering
        \includegraphics[width=0.8\linewidth]{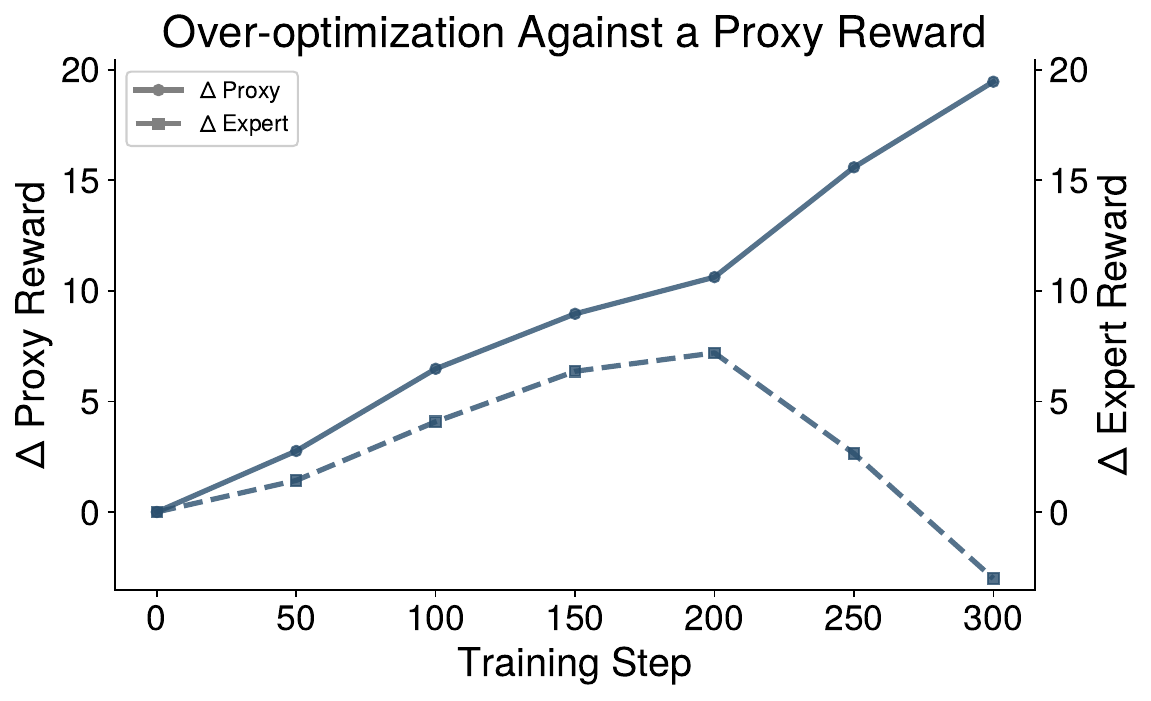}
        \caption{Haiku 4.5 setting.}
        \label{fig:baselines_haiku}
    \end{subfigure}
    \caption{RL training against the proxy reward model leads to an initial increase in expert reward, and over-optimization within a few hundred steps. The exact point of over-optimization can be highly trajectory-dependent.}
    \label{fig:baselines_combined}
\end{figure}

\textbf{Models}\hspace{3mm} We use Claude Haiku 4.5 as our policy and proxy reward model, and Claude Opus 4.5 (with up to 32K reasoning tokens) as our expert model \citep{anthropic2025haiku45, anthropic2025opus45}. We exclude the reasoning tokens from the natural language feedback provided to the proxy reward model.

\subsection{Baselines}

First, we establish the maximum possible performance and performance of weak model grading in our settings. All experiments use GRPO \citep{shao2024deepseekmathpushinglimitsmathematical} for RL training.

\begin{enumerate}
    \item \textbf{Maximum possible performance: direct training with expert rewards}. As the expert model produces both detailed natural language feedback and a scalar reward, we can just train against the scalar reward. This serves as a high water mark for performance on our tasks, but is not practically tractable for language model training. We find that training for up to $\mathcal{O}(10^5)$ samples leads to strong performance gains on both tasks. For creative writing and alignment research, expert reward increases by 39 points and 18 points respectively (see Figures \ref{fig:fig1_haiku} and \ref{fig:fig1}). Further training leads to mode collapse, and we manually truncate training based on the degree of collapse. We take the expert reward at this cut-off point to be the maximum achievable performance, and use it to define the performance gap.
    \item \textbf{Weak model grading baseline: direct training with proxy rewards}. We can also train directly against the proxy model, with the grading prompt created for the expert model. We find that this quickly leads to over-optimization (see Figure \ref{fig:baselines_combined}). As would be realistic for real-world training, we assume no knowledge of when to stop optimization. To measure the maximum performance of this baseline, we checkpoint and evaluate the policy for over-optimization every 50-100 RL steps using the expert model, stop once expert reward begins to decrease, and take the validation set reward from the previous checkpoint. Overall, RL against the weak proxy recovers 8\% and 39\% of the performance gap for our Qwen3-8B and Haiku 4.5 settings, respectively.
\end{enumerate}

We also test a reflective prompt optimization baseline \citep{agrawal2025gepareflectivepromptevolution}, using natural language feedback to iteratively update task prompts given an expert evaluation budget. We find that for a range ($\mathcal{O}(10^2) -\mathcal{O}(10^3)$) of label budgets, prompt optimization recovers no improvements in the Qwen3-8B setting and 25\% of the performance gap in the Haiku 4.5 setting, suggesting in-weight updates are necessary for maximum performance.

%% file: Sections/3_icl.tex

Our core aim is to recover more of the performance gap, while requiring fewer expert feedback samples, by efficiently leveraging online expert feedback. We aim to do this by distilling natural language feedback from the expert model into the proxy reward model.

As our proxy reward model is a post-trained language model, we may be able to sample-efficiently update the reward model by incorporating expert supervision directly into the prompt. In fact, when given the same prompt used for the expert model, the proxy model already produces rewards that are positively correlated with expert rewards (see Section \ref{sec:tracking_reward_alignment}); training against this baseline grader can already meaningfully increase expert reward, as in Figure \ref{fig:baselines_combined}.

To improve on this baseline, we increase the alignment and robustness of the proxy reward model by modifying grading prompts with few-shot examples or rubric-based structured guidance. Our main in-context learning method uses the following procedure, starting from the initial checkpoint:

\begin{enumerate}
    \item Sample fresh completions for train and validation prompts using the current checkpoint.
    \item Collect grading traces from the expert model on these completions, and add them to a \textbf{sample replay buffer}.
    \item Generate a prompt for the proxy model, using either rubrics (Section \ref{sec:rubrics}) or few-shot prompting (Section \ref{sec:fewshot}), incorporating the expert's natural language feedback on $N$ train samples. Then measure the proxy-expert grader alignment, computed as the prompt-averaged advantage correlation between proxy and expert rewards, on $M$ validation samples (pulling from the replay buffer)
    \item Repeat Step 3 $n_{\text{search}}$ times in parallel and ensemble the top $k$ highest-alignment prompts to form the proxy reward. There is substantial variation in alignment between different prompts (around $\pm 0.1$, see Figure \ref{fig:selection}), and number of validation samples required is independent of $n_{\text{search}}$, so we can get a better proxy reward using parallel search without needing additional expert samples.
    \item Train against the proxy reward for a fixed number of steps $S$, resampling and averaging over $n$ grading traces per completion. Resampling increases the proxy-expert reward alignment and reduces variance.
    \item Check for over-optimization, by collecting a small number $n_{\text{check}}$ of expert feedback samples to track whether the expert reward has decreased. We track over-optimization because we find it is difficult to reverse policy over-optimization once it has occurred.
    \item Add the expert samples used to track over-optimization to the sample replay buffer.
    \item If expert reward has not decreased significantly (i.e., by less than one standard deviation) from the last checkpoint, continue RL training from Step 5, else return to Step 1 to update the proxy grader prompt.
\end{enumerate}

By default, the prompts and rollouts we use to track over-optimization, update grader prompts, or measure correlation are selected at random. We vary $N$ depending on the prompting setup. For the Qwen3-8B setting we use $M = 800$ (50 validation prompts subsampled from the evaluation set, 16 samples each), $n_\text{search} = 20$, $k = 3$, $S = 50$, $n = 6$, and $n_\text{check} = 100$. For the Haiku 4.5 setting we use $M = 960$ (60 validation prompts, 16 samples each), $n_\text{search} = 10$, $k = 3$, $S = 50$, $n = 3$, and $n_\text{check} = 100$.

\subsection{Rubric prompts}
\label{sec:rubrics}

\begin{figure}[t]
    \centering
    \begin{subfigure}[t]{\linewidth}
        \centering
        \includegraphics[width=0.7\linewidth]{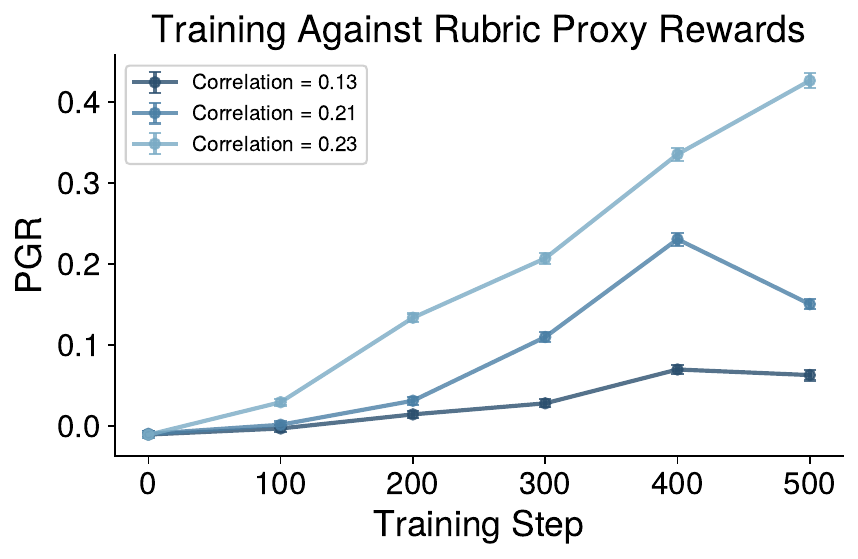}
        \caption{Rubric-based proxy reward models.}
        \label{fig:rubric_comparison}
    \end{subfigure}
    \\[1ex]
    \begin{subfigure}[t]{\linewidth}
        \centering
        \includegraphics[width=0.7\linewidth]{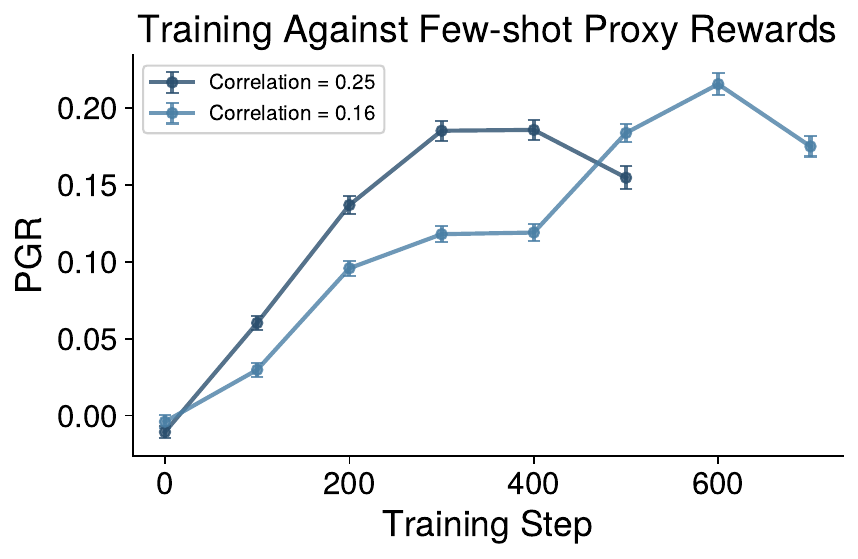}
        \caption{Few-shot prompted proxy reward models.}
        \label{fig:fewshot_comparison}
    \end{subfigure}
    \caption{(Qwen3-8B setting) RL training against proxy reward models with differing initial advantage alignment. Initial reward alignment can be predictive of downstream RL performance.}
    \label{fig:proxy_comparison}
\end{figure}

The first prompting method we use to improve the proxy model's grading ability is rubrics, which provide explicit guidance on how to weight the correct qualitative factors and assign grades. We generate rubrics by providing a strong language model -- the same language model that is standing in for a human expert -- a set of completions from the policy, grading traces from the expert model, and grading traces from the proxy reward model, with $N = 10$ (see Appendix \ref{sec:prompts} for a generation prompt). This is analogous to the human expert hand-writing a rubric. Rubrics with higher alignment on the initial distribution usually lead to higher maximum PGR, as shown in Figure \ref{fig:rubric_comparison}.

\begin{figure*}[htbp]
    \centering
    \begin{subfigure}[t]{\linewidth}
        \centering
        \includegraphics[width=\linewidth]{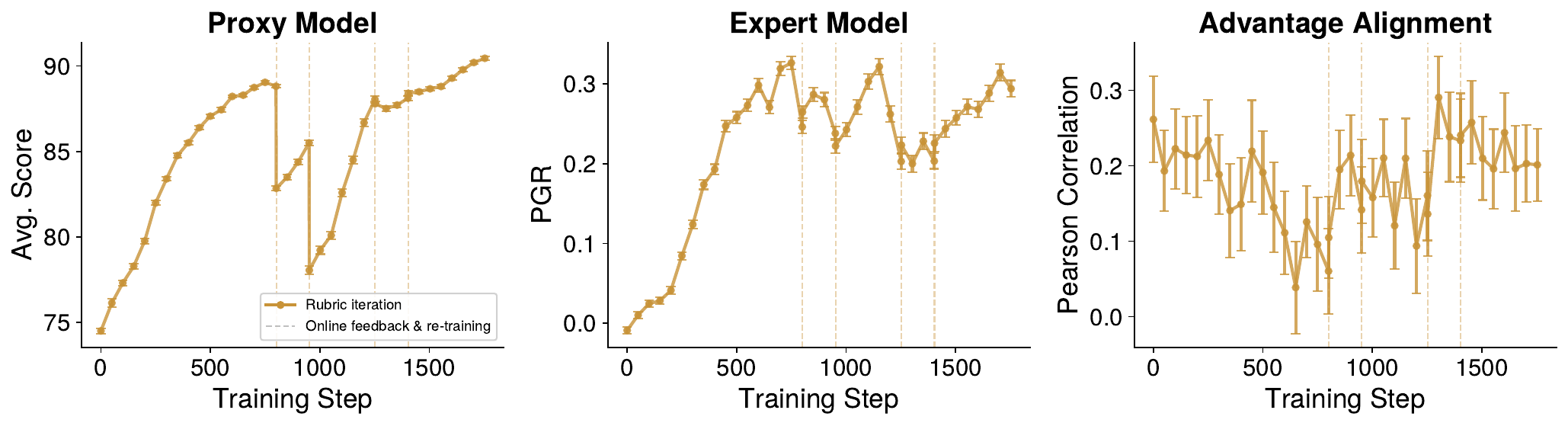}
        \caption{(Qwen3-8B, rubric grading) While in the first iteration the proxy reward is useful, in successive iterations rubric updates do not lead to further expert reward increases; proxy reward models with reasonable advantage alignment still over-optimize quickly. In later iterations the proxy reward is nearly saturated even after online feedback.}
        \label{fig:rubric_iteration}
    \end{subfigure}

    \vspace{0.5em}

    \begin{subfigure}[t]{\linewidth}
        \centering
        \includegraphics[width=\linewidth]{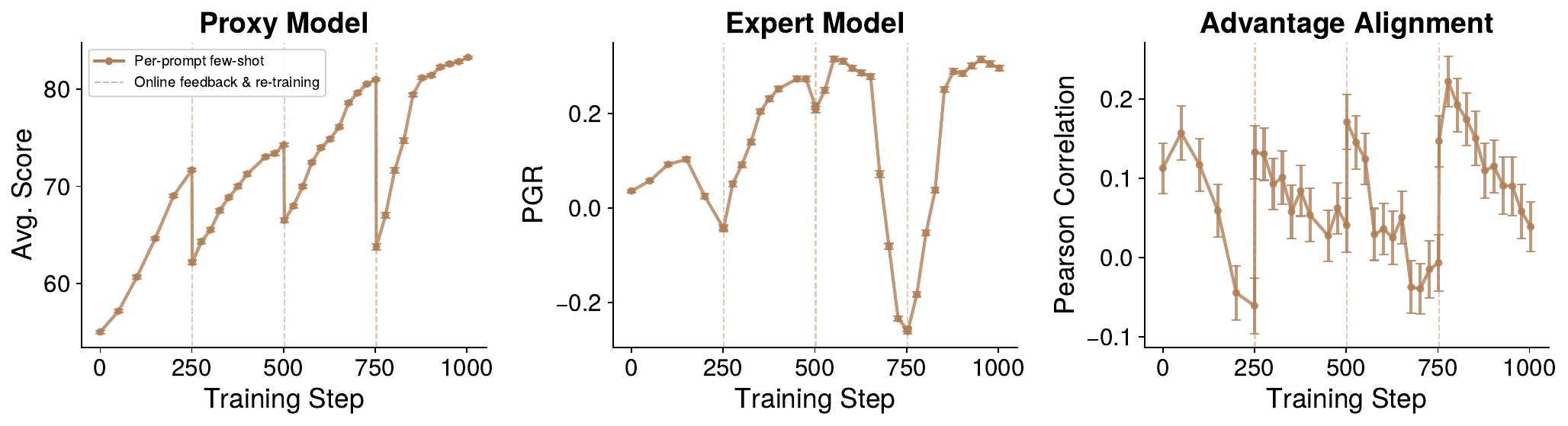}
        \caption{(Qwen3-8B, per-task few-shot prompt, 2 examples per prompt) Expert reward plateaus after a couple of iterations. The sharp decrease in expert reward at step 700 is caused by exploding lengths and language switching.}
        \label{fig:perprompt_fs_iteration}
    \end{subfigure}

    \vspace{0.5em}

    \begin{subfigure}[t]{\linewidth}
        \centering
        \includegraphics[width=\linewidth]{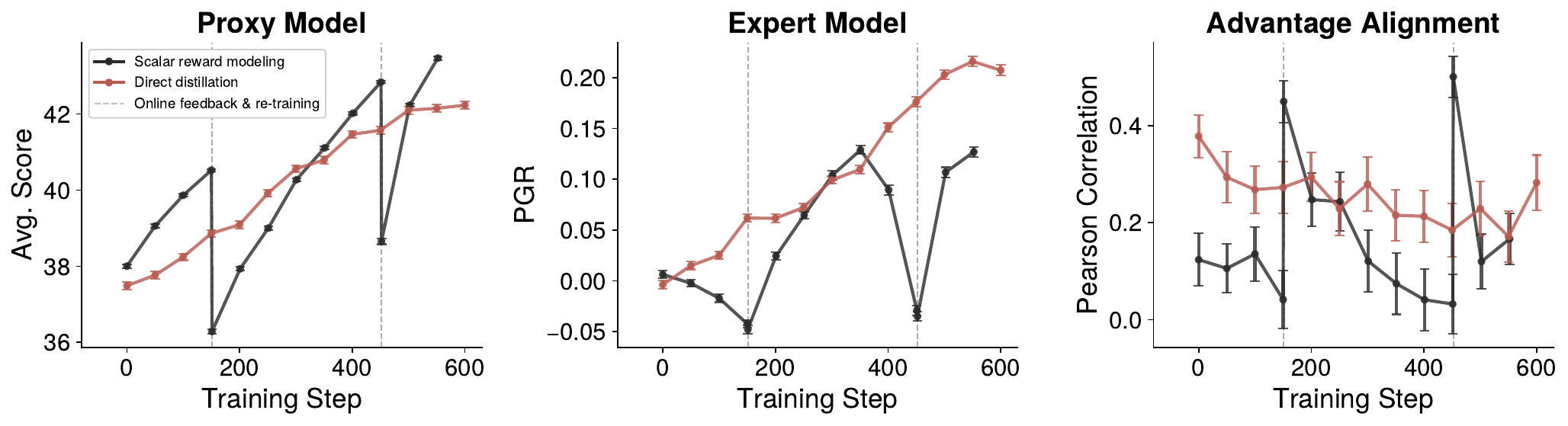}
        \caption{(Qwen3-8B, full-trace distillation versus scalar reward only) The scalar-only proxy reward over-optimizes more quickly and the reward alignment decreases rapidly.}
        \label{fig:rm_vs_distillation}
    \end{subfigure}
    \caption{Example training runs from the Qwen3-8B setting. Figures show the proxy and expert reward, plus correlation between expert and proxy advantages over the course of RL training, with iterative grader realignment. We realign the grader using different methods, as described in Sections \ref{sec:icl} and \ref{sec:ft}.}
    \label{fig:icl_combined}
\end{figure*}

In the Qwen3-8B setting, we train with rubrics for 4 total iterations of grader realignment, and recover up to 35\% of the performance gap; the evolution of the proxy reward, expert reward and grader advantage alignment are shown in Figure \ref{fig:rubric_iteration}. We use around 3K total expert feedback samples by the end of the run, or roughly 50x fewer samples than the strong baseline. Performance does not increase substantially after the first grader realignment, and after the first realignment rubrics tend to saturate more easily. Our interpretation is that, after optimizing against the proxy reward, it becomes increasingly difficult to adjust the proxy's judgment via prompt changes.

In the Haiku 4.5 setting, we train with rubrics for 3 iterations, and recover about 35\% of the performance gap using about 30x fewer samples. Performance does not substantially increase after the first iteration. For Haiku 4.5, rubric iteration also does not out-perform training against the proxy model with a baseline prompt.

\subsection{Few-shot prompts}
\label{sec:fewshot}

We also attempt to improve the proxy reward by providing full grading examples from the expert model as a few-shot prompt ($N = 5$). We test the following setups:
\begin{itemize}
    \item \textbf{In-prompt examples}: We include few-shot examples as part of a single \texttt{user} prompt.
    \item \textbf{Pre-filled examples}: We pre-fill (prompt, full grading trace) pairs from the expert model into the \texttt{user} and \texttt{assistant} turns.
\end{itemize}

In both the Qwen3-8B and Haiku 4.5 settings, using either prompt structure, few-shot prompts are largely ineffective. While we can find prompts with high expert-proxy reward alignment, we over-optimize against few-shot prompts more quickly than rubrics (see Figure \ref{fig:fewshot_comparison}). When re-aligning the proxy reward model after the first iteration of optimization, few-shot prompts are generally ineffective at meaningfully increasing reward alignment.

\begin{figure*}
    \centering
    \includegraphics[width=0.48\linewidth]{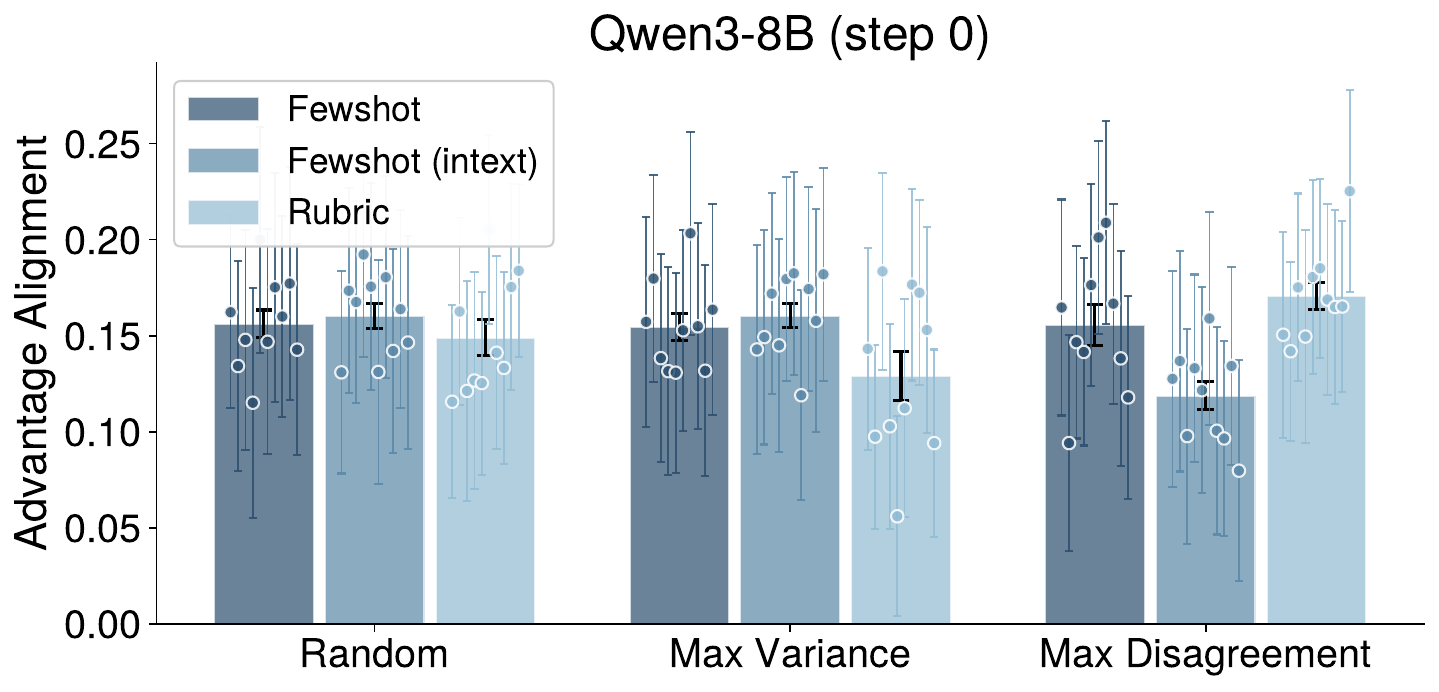}
    \hfill
    \includegraphics[width=0.48\linewidth]{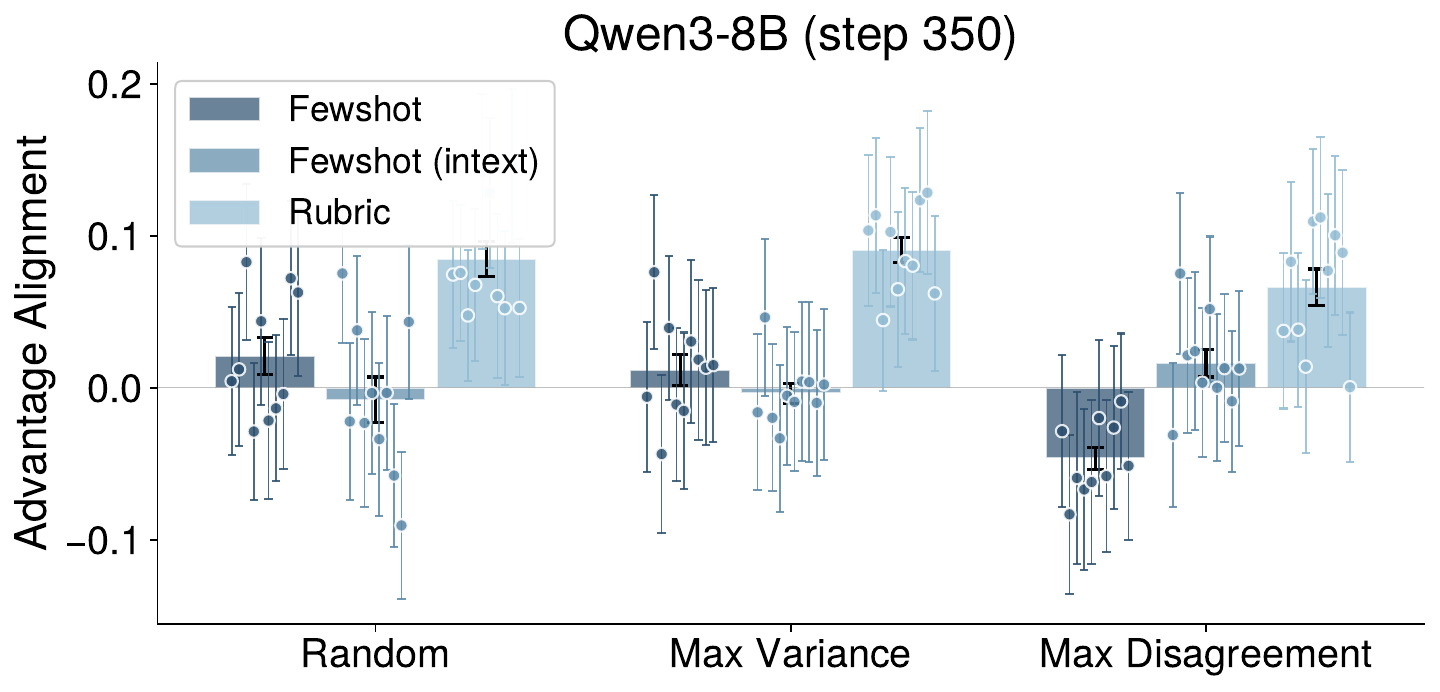}
    \caption{(Qwen3-8B setting) Reward alignment for proxy reward models aligned using various in-context learning methods, computed at step 0 and step 350 (after optimizing against the initial proxy reward, then collecting online feedback and re-aligning). After optimization, even after online feedback and realigning, the proxy reward model generally does not match the reward alignment at step 0.}
    \label{fig:selection}
\end{figure*}

\subsection{Per-task prompts}
Our rubric and few-shot experiments use a single grading prompt for all tasks in a setting. We also explore generating grading prompts for each task. We sweep over [2, 4, 6] expert feedback samples for each grading prompt (either a few-shot prompt or a natural language rubric). In these experiments we do not track over-optimization or perform parallel search over prompts.

In the Qwen3-8B setting, we run 4 iterations for both per-prompt rubrics and per-prompt few-shot examples. We see minimal expert reward increase with per-prompt rubrics but recover around 35\% of the performance gap using per-prompt few-shot examples while requiring around 5K expert samples (results using 2 samples per prompt shown in Figure \ref{fig:perprompt_fs_iteration}). On each iteration, updating the per-prompt few-shot examples also increases reward alignment.

In the Haiku 4.5 setting, we do not see any improvements from using per-prompt rubrics or few-shot examples (see Figure \ref{fig:haiku_combined} in the Appendix). 

\subsection{Selecting informative samples for online feedback}

How we select which samples get online feedback collection could in principle affect the quality of the resulting proxy reward model. For example, we might want to preferentially select samples for feedback that are difficult for the proxy reward model. We test the following feedback protocols:

\begin{itemize}
    \item \textbf{Random selection}: The samples included in-context are sampled at random from the current policy.
    \item \textbf{Max-variance selection}: For each sample $s$ we collect a set $G(s)$ of 10 proxy grades. We rank the samples based on $\frac{\text{stdev}(G(s))}{\text{mean}(G(s))}$ and pick the top $N$ for feedback.
    \item \textbf{Max-disagreement selection}: For each sample $s$, we calculate its advantage relative to its GRPO group from the RL run, using the proxy reward and ground truth reward, say $A_{\text{pr}}(s)$ and $A_{\text{exp}}(s)$ respectively. We then rank the samples based on $|A_{\text{pr}}(s) - A_{\text{exp}}(s)|$. This approach requires sub-sampling, as we must collect expert feedback for all samples to be ranked; the selection is still useful for methods like rubric writing or few-shot prompting, which can only directly absorb 5-10 samples of complete expert feedback.
\end{itemize}

We select from a sub-sampled pool of around 100 train samples. We find that the method of selecting samples does not improve proxy reward alignment on either the initial distribution, or when re-aligning the proxy reward on a later distribution after optimization, as in Figure \ref{fig:selection}. The difference in alignment between two prompts distilled from the same set of samples is larger than the difference in mean alignment between selection protocols.

%% file: Sections/4_finetuning.tex
Next, we consider methods which align the proxy reward model with the expert model by using supervised learning. Our approach is to distill the expert model's judgment into the proxy reward model, via multiple iterations of fine-tuning on online feedback collected by the expert. We use the following procedure:
\begin{enumerate}
    \item Initialize the proxy reward model by training on expert model grades of samples from the base policy.
    \item At a fixed interval (e.g. every few hundred steps):
    \begin{enumerate}
        \item Sample completions for all train prompts from the latest checkpoint.
        \item Collect fresh, on-policy grading traces from the expert model.
        \item Fine-tune the proxy reward model on the online feedback from the expert.
        \item Continue training from the current policy with the updated proxy reward model.
    \end{enumerate}
\end{enumerate} 

In contrast to our ICL experiments, where we track over-optimization to determine when to update the proxy reward model (Section \ref{sec:icl}), proxy reward models produced with SFT are generally more robust. Our experiments update the proxy reward at fixed step intervals, and were generally robust to reasonable interval sizes. We provide training details and hyperparameters in Appendix \ref{appendix:training}.


\subsection{Full-trace distillation}
\label{sec:dd}

We first consider distilling complete grading traces from the expert model, teaching the proxy reward model to fully replicate the reward assignment process. In this setting, fine-tuning tends to initially hurt the alignment between the expert model and the proxy reward model, thus requiring a sample-heavy initialization around $\mathcal{O}(10^3) - \mathcal{O}(10^4)$ samples; see Appendix \ref{appendix:sft} for details. 

In the Qwen3-8B setting, we test three sets of protocol hyperparameters: 
\begin{itemize}
    \item \textbf{Heavy}: 20K initial samples, 800 RL steps per iteration before grader realignment, 10K additional samples per iteration, 3 epochs of SFT
    \item \textbf{Light}: 9K initial samples, 500 steps per iteration, 4.5K additional samples per iteration, 3 epochs of SFT
    \item \textbf{Ultra-light}: 3K initial samples, 500 steps per iteration, 1.5K additional samples per iteration, 9 epochs of SFT
\end{itemize}
We find full-trace distillation is highly effective. With \textbf{heavy} settings, we can recover the full performance gap. With \textbf{light} and \textbf{ultra-light} settings, we can recover around 80\% of the performance gap. Scaling is slightly less effective in the lighter settings primarily due to training instability that emerges in later iterations. While we fix RL hyperparameters for parity across experiments, tuning hyperparameters could potentially prevent this instability.

We find qualitatively similar results in the Haiku 4.5 setting (see Figures \ref{fig:fig1_haiku} and \ref{fig:haiku_combined}). In both \textbf{heavy} (6K initial samples, 250 steps per iteration, 3K additional samples) and \textbf{light} (3K initial samples, 250 steps per iteration, 1.5K additional samples) settings, we can fully recover the performance gap.

\subsection{Context distillation}
Rather than training on full expert grader traces, we can use context distillation. Given a completion from the policy and a grading trace from the expert grader, we condition the proxy model on additional information from the expert grader. We then generate proxy grader traces, filter out ones which mention the additional information and fine-tune the proxy reward model on the traces. This reduces distributional differences when fine-tuning and allows us to generate different ``rewrites'' of the same grading trace. 

We consider two approaches to context distillation:
\begin{itemize}
    \item \textbf{Full traces}: We include the grader's feedback, score deliberation, and scalar score in the prompt for the proxy model.
    \item \textbf{Feedback only}: We include only the grader's natural language feedback. This does not require the expert to provide scalar rewards, which may be a more realistic analogy for human experts providing feedback.
\end{itemize}

We train on up to 3 grading traces per completion. In the Qwen3-8B setting, we train with 3K initial samples, 500 steps per iteration, and 1.5K additional samples per iteration, 3 epochs. This is the same volume of data as the ``ultra-light'' setting in Section \ref{sec:dd}.

We find that both context distillation methods perform equally well; in other words, the proxy reward model learns as much from \textit{just the feedback} as the full grader traces. Context distillation performs as well as data-matched full-trace distillation; both recover approximately 80\% of the performance gap, and we truncate training when it becomes unstable. 

\subsection{Scalar reward modeling}
We can also train the proxy reward model with an additional scalar head to directly predict the expert model's scalar scores, discarding the expert's natural language feedback. In the Qwen3-8B setting, we train a scalar reward matching the data volume of heavy full-trace distillation.

The scalar-only proxy reward model is significantly less robust to optimization pressure than the proxy reward model trained with full distillation, with over-optimization occurring within 200 steps of RL on every iteration, as shown in Figure \ref{fig:rm_vs_distillation}. While the online updates do successfully re-align the proxy reward, we still over-optimize rapidly against the updated reward.

%% file: Sections/5_alignment.tex
\subsection{Estimating expert reward movement} \label{sec:estimators}
Can we tell, without actually doing RL training, how robust a proxy reward is? We observe that better proxy reward models generally have higher and more robust agreement with expert rewards.

Recall our goal is to find a policy $\pi_\theta$ which maximizes the expected expert reward ($R_{\text{exp}}$) on the task distribution $D$:
$$
J_{\text{exp}}(\theta) = \mathbb{E}_{x \sim D, y \sim \pi_{\theta}(\cdot | x)}[R_{\text{exp}}(x, y)]
$$ 
We attempt to increase $J_{\text{exp}}$ by doing RL against proxy rewards $R_{\text{pr}}$. Let $\Delta \theta_t$ be the weight update from a single step of RL against $R_{\text{pr}}$. Then to first order, $\Delta J_{\text{exp}}$ is the projection of the directional derivative of $J_{\text{exp}}$ onto the actual update:
$$
\Delta J_{\text{exp}} \approx \nabla J_{\text{exp}}(\theta_t)^{\top} \Delta \theta_t,
$$
where we can compute $\nabla J_{\text{exp}}(\theta_t)$ by simply swapping the proxy rewards for expert rewards in the gradient. If we are using vanilla gradient descent, then $\Delta \theta_t \propto \nabla J_{\text{pr}}(\theta_t)$. We can estimate this on each step of RL. 

To estimate proxy reward model quality independent of the training setup or without access to gradients, note that the ground truth reward changes with changes in trajectory probabilities via:
$$\Delta J_{\text{exp}} = \mathbb{E}_{x \sim D} \left[ \sum_y \Delta \pi(y \mid x) R_{\text{exp}}(x, y) \right]$$

Approximating RL as independently up- or down-weighting trajectory probabilities proportional to the computed proxy advantage:
$$
\Delta \pi(y \mid x) \approx \eta \pi(y \mid x)A_\text{pr}(x, y),
$$
we get the expression
$$
\Delta J_{\text{exp}} \approx \eta \mathbb{E}_{x \sim D, y \sim \pi(\cdot \mid x)} [R_{\text{exp}} A_{\text{pr}}],
$$
which is the prompt-averaged inner product between proxy and ground truth advantages. Note, for variance reduction we can do per-prompt mean-centering of the expert rewards.

\begin{figure}[tb]
    \centering
    \begin{subfigure}[t]{\linewidth}
        \centering
        \includegraphics[width=0.75\linewidth]{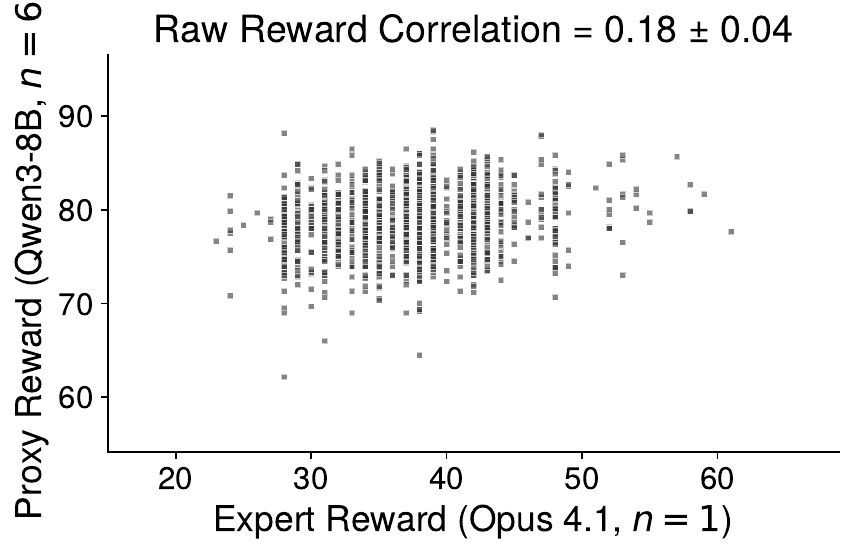}
        \caption{Qwen3-8B setting.}
        \label{fig:scatter}
    \end{subfigure}
    \\[1ex]
    \begin{subfigure}[t]{\linewidth}
        \centering
        \includegraphics[width=0.75\linewidth]{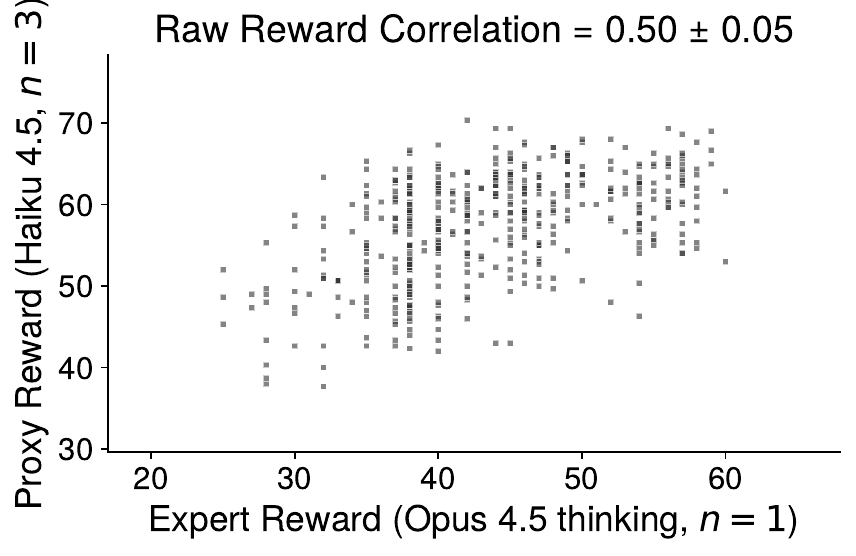}
        \caption{Haiku 4.5 setting.}
        \label{fig:scatter_haiku}
    \end{subfigure}
    \caption{Proxy vs.\ expert reward, using the baseline grading prompt and re-sampling the proxy reward to reduce variance.}
    \label{fig:scatter_combined}
\end{figure}

Under these simplifying assumptions, if the expert and proxy rewards are positively correlated on a given prompt distribution, we should see increases in the expert reward by training against the proxy reward. In practice, with the default grading prompts we do observe positive correlations between proxy and expert rewards, as shown in Figure \ref{fig:scatter_combined}. This motivates two reward alignment statistics:
\begin{itemize}
    \item Prompt-averaged advantage Pearson correlation: \begin{equation*}
    \rho = \mathbb{E}_{x \sim D}\left[ \frac{\mathbb{E}_y[A_\text{exp}A_\text{pr} \mid x]}{\sigma[A_{\text{exp}} \mid x] \sigma[A_{\text{pr}} \mid x]} \right]
\end{equation*}

     \item Raw reward Pearson correlation, aggregated across prompts. (Correlation between raw rewards may be biased if one reward model has systematic preferences towards certain groups.)
\end{itemize}

We estimate all uncertainties by hierarchical bootstrap (over prompts and rollouts), and find that $\mathcal{O}(10^3)$ samples are required to cleanly measure distribution-level correlation.

\subsection{Tracking reward alignment during RL}
\label{sec:tracking_reward_alignment}

Throughout this work, we have measured the advantage correlation as a proxy for the reward alignment between the proxy and expert judgments. In general, tracking reward alignment over the course of an RL run provides useful context on runs. For example, the difference in effectiveness between scalar reward modeling and direct distillation (Figure \ref{fig:rm_vs_distillation}) can be understood by tracking reward alignment: each update sharply increases alignment on the current policy's distribution, but because the scalar reward model is less robust, the expert reward and advantage alignment sharply decline after a few hundred steps. In comparison, the proxy trained on full traces is much more robust.

The following heuristics can be useful for understanding and improving online feedback protocols:
\begin{enumerate}
    \item If there is a positive correlation between the proxy and experts, training against the proxy reward sometimes \textit{but not always} leads to increases in the expert rewards. See Figure \ref{fig:rubric_iteration}, where on later iterations the alignment is sometimes positive while the expert reward still decreases.
    \item Given a choice between proxy reward models created using similar procedures (e.g. rubric generation, few-shot prompting), it is better to choose proxy reward models with higher alignment (see Figure \ref{fig:rubric_comparison}).
    \item The alignment between the proxy and expert rewards decreases during RL optimization. If it is easier to over-optimize against a given proxy reward, the alignment also tends to decrease faster (Figures \ref{fig:sft_alignment}, \ref{fig:iterative}).
\end{enumerate}

%% file: Sections/6_related_work.tex
\subsection{RLHF and Over-optimization}
Reinforcement learning from human feedback is a standard method for aligning language models with human preferences on diverse and subjective tasks \citep{christiano2017deep, ziegler2019finetuning, stiennon2020learning, ouyang2022training}. Traditionally, human preferences are collected as pair-wise rankings, which can then be distilled into a scalar reward model, or trained against directly \citep{rafailov2023direct, ji2025reinforcementlearninghumanfeedback}. Training against such reward models initially improves model performance according to human preferences, but eventually leads to over-optimization \citep{gao2022scalinglawsrewardmodel, wen2024languagemodelslearnmislead, sharma2025understandingsycophancylanguagemodels, rafailov2024scalinglawsrewardmodel}. Several directions address this by enriching the reward signal, e.g., fine-grained reward models \citep{wu2023finegrained}, or by increasing reward model robustness through ensembling \citep{coste2024rewardmodelensembleshelp} or constrained optimization \citep{moskovitz2023confrontingrewardmodeloveroptimization}.

\subsection{Learning from Natural Language Feedback}
By default, reinforcement learning requires scalar rewards or pair-wise preferences, which contain much less information than long-form textual feedback. A line of work exploits this richer signal at inference time: gradient-free techniques that use detailed natural language feedback can improve or adapt language model systems more sample-efficiently than weight updates \citep{yang2024largelanguagemodelsoptimizers}. For example, prompt optimization methods such as GEPA, which uses natural language reflection \citep{agrawal2025gepareflectivepromptevolution}, and MIPRO \citep{opsahlong2024optimizinginstructionsdemonstrationsmultistage} can sometimes out-perform in-weight updates in substantially fewer iterations. Other inference-time methods for improving language model system performance using textual feedback include TextGrad \citep{yuksekgonul2024textgradautomaticdifferentiationtext} and Feedback Descent \citep{lee2025feedbackdescentopenendedtext}.

\subsection{Scaling Human Supervision}
Beyond RLHF, several lines of work scale human supervision by using language models to interpret natural language and generate training signal. Constitutional AI \citep{bai2022constitutional} uses language models to generate fine-tuning and preference data from a human-written list of natural language principles; \citet{scheurer2022training} similarly condition on natural language feedback to generate improved completions, which they then distill. Other methods rely on human-written or model-written rubrics to improve the quality of model judgment and scalar reward assignment \citep{gunjal2025rubricsrewardsreinforcementlearning, kimiteam2025kimik2openagentic, sharma2025researchrubricsbenchmarkpromptsrubrics, Hashemi_2024, zhang2025chasingtaileffectiverubricbased}.

A complementary line of work trains models to serve as evaluators or critics directly. This has been done by distilling judgments from stronger models \citep{kim2023prometheus} or from humans \citep{jin2023dataefficientalignmentlargelanguage}, by training generative reward models that reason before scoring \citep{mahan2024generativerewardmodels}, and by using human supervision to improve model-written critiques \citep{mcaleese2024llmcriticshelpcatch}.

%% file: Sections/7_discussion.tex
Training language models on fuzzy tasks will require expert human supervision which, as model capabilities advance and model outputs become more complex, will be increasingly expensive to collect. In this work, we study whether natural language expert feedback can improve the data efficiency of expert supervision, through protocols that distill online feedback into proxy reward models via in-context learning or fine-tuning. We find that online in-context learning methods only recover around a third of the performance gap, but can improve data efficiency by up to 50x, and that online fine-tuning methods can fully or nearly recover the performance gap while improving data efficiency up to 20x.

We find similar qualitative trends in both settings studied. Our results from eliciting Qwen3-8B suggest that learning online from expert feedback can improve both the alignment and robustness of proxy reward models across multiple iterations, for both in-context learning and supervised fine-tuning methods. When scaling up to Haiku 4.5, our results suggest ICL-based online feedback methods generally do not outperform the prompted proxy model baseline, which is much stronger than in the Qwen3-8B setting; SFT-based methods remain highly effective. This suggests that in-weight updates from feedback are most effective, but that more intelligent models might benefit less from explicit reward assignment guidance or scaffolding.

Finally, we discuss some limitations of our study:
\begin{itemize}
    \item \textbf{Realism}: We study the problem of scaling expert oversight using a scaled-down analogy, with a language model standing in as the ``expert''. While the expert model's long-form feedback and reward assignment are quite variable, this is a simplification compared to actual expert preferences.
    \item \textbf{Scope}: Due to the cost of full RL training, we study only two tasks and two sets of models.
    \item \textbf{Randomness}: We only run 1-2 seeds for each method. While the between-run variation is small relative to the performance differences between methods, there is still significant variation between RL runs with identical data but different random seeds (see Figure \ref{fig:baselines}).
\end{itemize}

%% file: Sections/8_appendix.tex
\section{Training Details} \label{appendix:training}

\subsection{Qwen3-8B Setting (Creative Writing)}
Unless otherwise specified, we train all parameters.

\paragraph{RL Algorithm} We use a policy gradient-based algorithm with group-based advantage normalization and token-level normalization, closely resembling GRPO and DAPO \cite{shao2024deepseekmathpushinglimitsmathematical, yu2025dapoopensourcellmreinforcement}. We train fully on-policy, using importance sampling to adjust for numerical differences between inference and training kernels. We use constant learning rate of 7E-7 with no warmup, no KL penalty, Adam optimizer, a group size of 8 rollouts, a batch size of 8 prompts, gradient clipping to norm 1.0, and epsilon clipping to 0.2. We train with a maximum output length of 5,000 tokens.

\paragraph{Supervised Fine-Tuning} We train directly on SFT samples with next-token prediction. Unless otherwise specified, we fine-tune for 3 epochs with a learning rate of 1E-5, 10\% warmup steps, linear learning rate decay, AdamW optimizer with no weight decay, and a batch size of 16 sequences.

\paragraph{Scalar Reward Modeling} We train a scalar output head on the final hidden state, using the Huber loss. All hyperparameters match those of SFT.

\subsection{Haiku 4.5 Setting (Alignment Research)}
We train with a maximum output length of 3,000 tokens. All proxy model grading is done with a maximum output length of 16,000 tokens; expert model grading is done with up to 32,000 reasoning tokens and 32,000 output tokens.

\section{Additional Results}

\subsection{Sample Efficiency of Distillation} \label{appendix:sft}

\begin{figure}[h]
    \centering
    \includegraphics[width=0.48\textwidth]{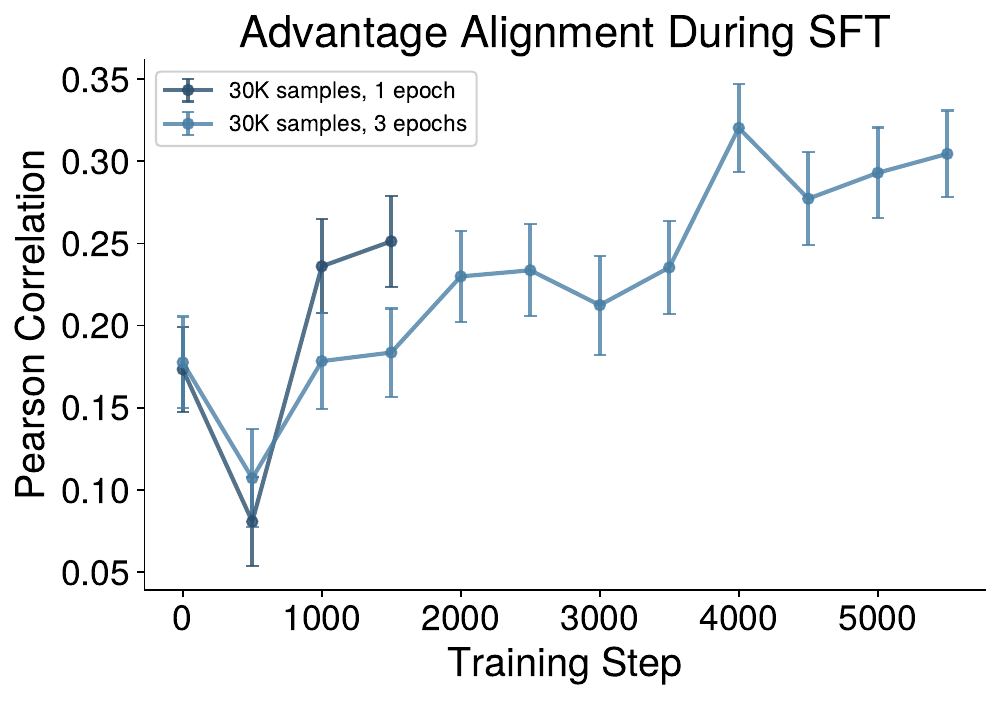}
    \caption{(Qwen3-8B setting) The correlation between proxy and expert advantages first decreases, then increases, during full-trace distillation.}
    \label{fig:sft_alignment}
\end{figure}

In the Qwen3-8B setting, performing supervised fine-tuning on a small number of samples actually hurts proxy-expert reward alignment relative to this baseline. This can be fixed by training with a larger number of samples and over multiple epochs, after which the trained reward model has substantially higher alignment with the expert; see Figure \ref{fig:sft_alignment}.  To address this we recommend training the proxy model on a large initial batch of expert samples.

\begin{figure*}[htbp]
    \centering
    \begin{subfigure}[t]{\linewidth}
        \centering
        \includegraphics[width=\linewidth]{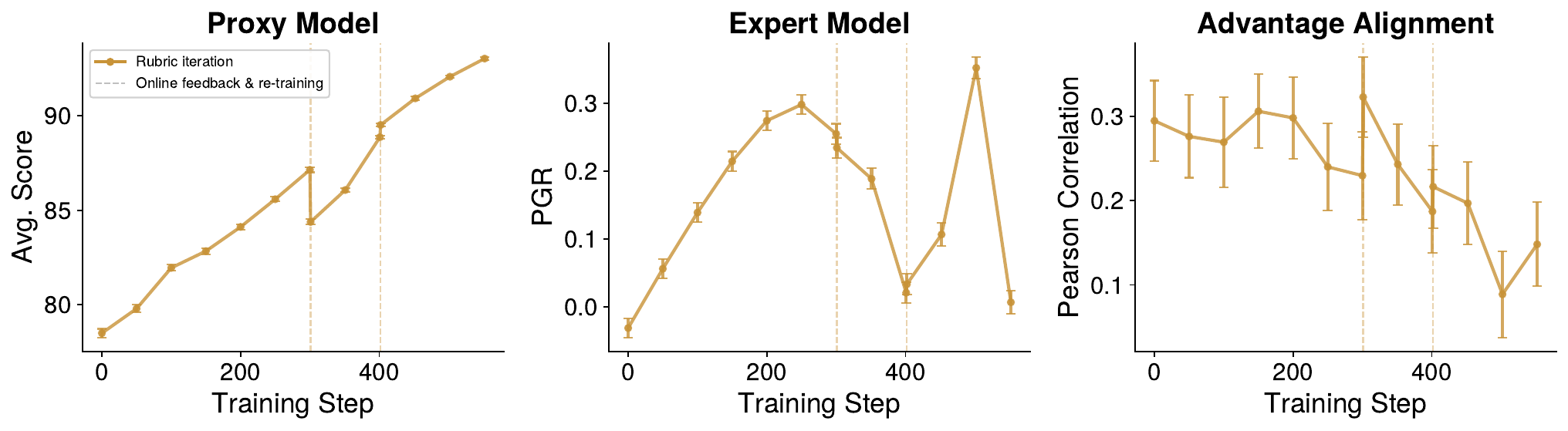}
        \caption{RL training with proxy reward models updated via iterative rubric writing.}
        \label{fig:rubric_iteration_haiku}
    \end{subfigure}

    \vspace{0.5em}

    \begin{subfigure}[t]{\linewidth}
        \centering
        \includegraphics[width=\linewidth]{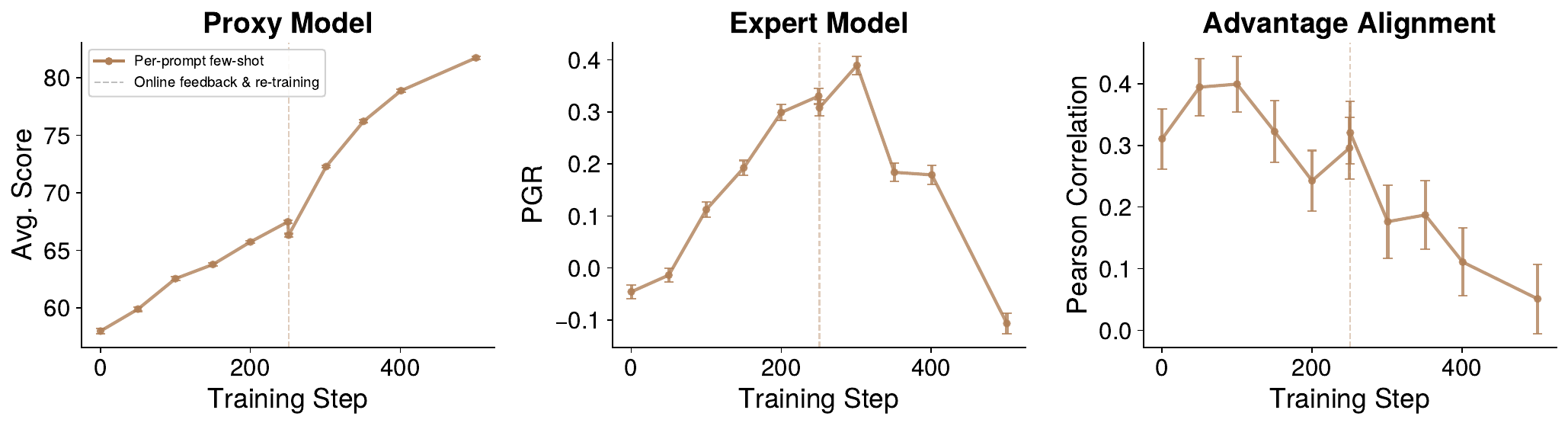}
        \caption{RL training with proxy reward models updated via per-prompt few-shot examples.}
        \label{fig:perprompt_fs_haiku}
    \end{subfigure}

    \vspace{0.5em}

    \begin{subfigure}[t]{\linewidth}
        \centering
        \includegraphics[width=\linewidth]{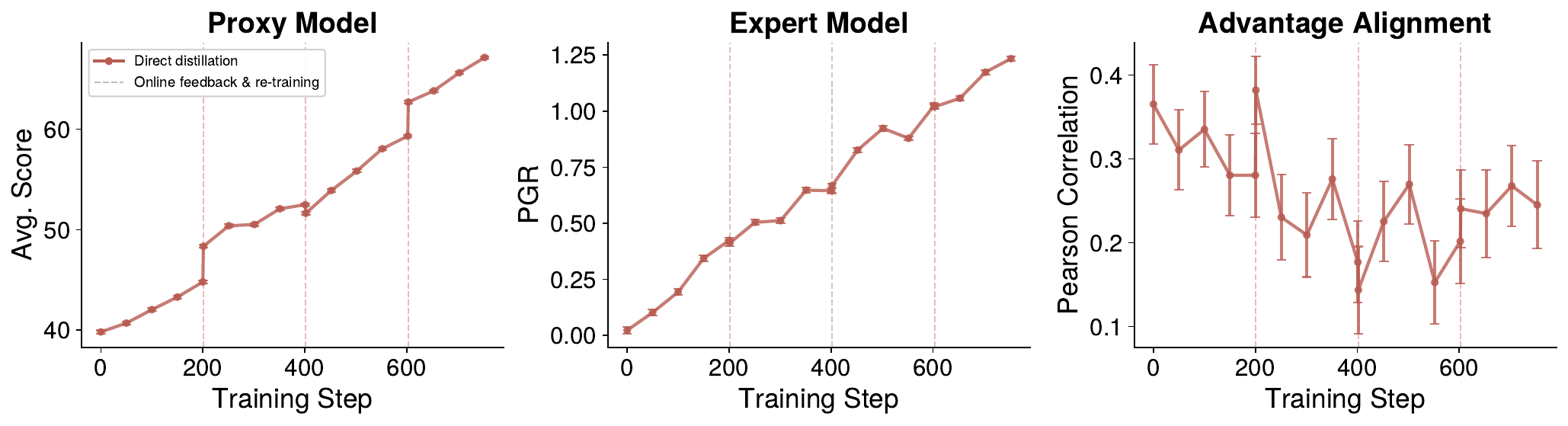}
        \caption{RL training with proxy reward models updated via direct distillation, using the ``light'' configuration.}
        \label{fig:distillation_haiku}
    \end{subfigure}
    \caption{Haiku 4.5 (alignment research) setting. Compared to the Qwen3-8B setting (Figure \ref{fig:icl_combined}), we observe more dramatic and sudden over-optimization; additional rubric updates, few-shot examples, or distillation iterations often did not meaningfully increase the reward alignment.}
    \label{fig:haiku_combined}
\end{figure*}



\subsection{Re-training from Scratch} \label{appendix:iterative}
Instead of continuing training from the last checkpoint after collecting fresh feedback, we test training from scratch with the updated proxy reward model. We use the proxy reward models from each iteration of the full-trace distillation experiments in Section \ref{sec:ft}.

\begin{figure*}[h]
    \centering
    \includegraphics[width=\textwidth]{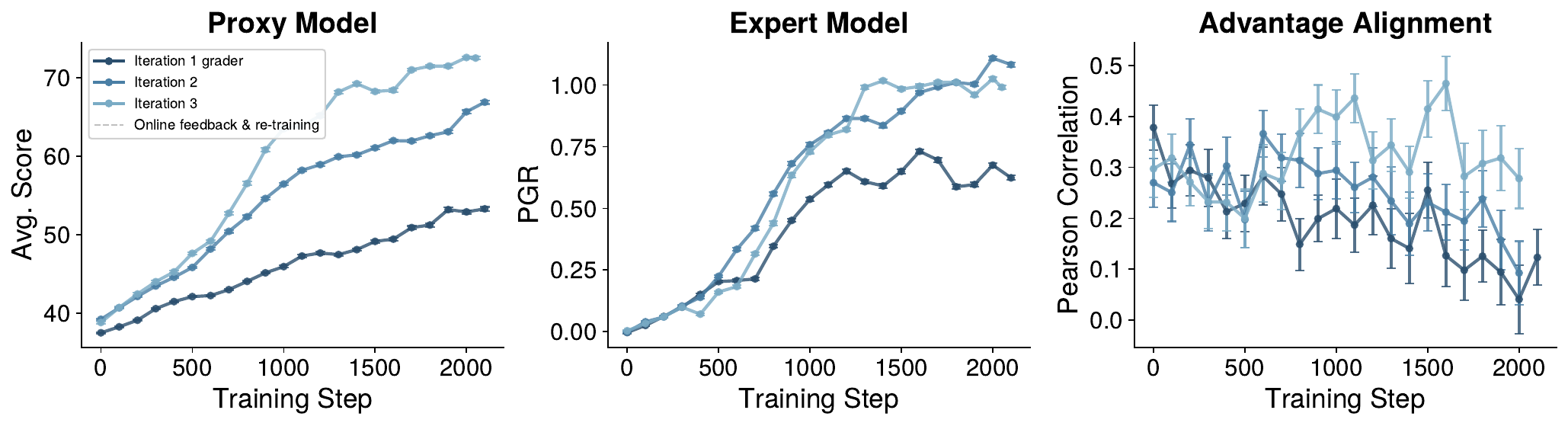}
    \caption{(Qwen3-8B setting) Training from scratch with proxy reward models from different iterations of the original online feedback protocol.}
    \label{fig:iterative}
\end{figure*}

Proxy reward models from all iterations of the online feedback protocol tend to have similar alignment with the expert grader when close to base policy. However, after several hundred steps of optimization, the proxy reward models from later iterations have consistently higher grade alignment. We are able to fully recover the performance gap when training from-scratch with proxy models from iterations 2 and 3, but not with the proxy model from iteration 1, which has only been supervised on completions from the base policy. See Figure \ref{fig:iterative}.

\subsection{Limitations of First-Order Approximations}
\begin{figure}[h]
    \centering
    \includegraphics[width=0.5\columnwidth]{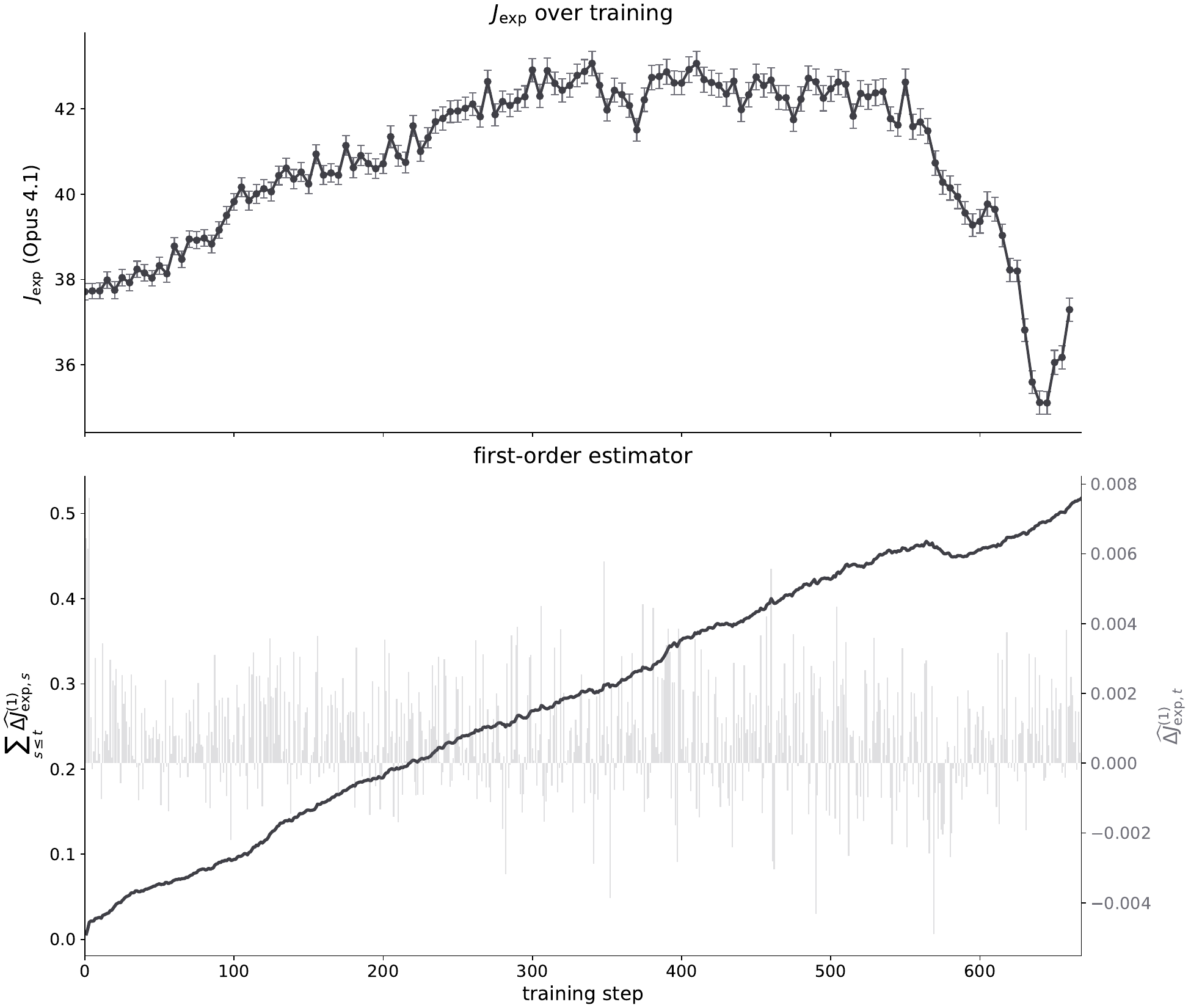}
    \caption{(Qwen3-8B setting) At e.g. step 560, the gradient-based first-order estimator (bottom) also predicts a decrease in expert reward, but much smaller in magnitude than the observed decrease (top)}
    \label{fig:first_order}
\end{figure}

In Section \ref{sec:estimators} we discussed how to estimate the expert reward increase from training against the proxy reward, and in Section \ref{sec:alignment} we offered some qualitative observations on this. In general, we find that our first-order approximations tend to \textit{overestimate} the increase in expert reward. In Figure \ref{fig:first_order}, we compare the actual observed expert reward trajectory with the cumulative trajectory predicted by the first-order approximation. We find that we overwhelmingly predict positive movement even when the empirical expert reward is decreasing. 

One possible reason is that there is generally asymmetry between reward increases and reward decreases in our settings. In particular, there are several pervasive issues, such as general incoherence, hallucinations, or overly lengthy completions, which can lead to very sudden reward collapse.


%% file: Sections/9_prompts.tex
\begin{figure*}[!ht]
\begin{promptbox}[Creative Writing: evaluation prompt]
\small
You are an expert creative writing evaluator and grader. You always provide detailed, precise, expert-level critiques of creative writing. You are also an accurate and considerate numeric grader.

You believe that a great story MUST have logical coherence and a strong technical foundation; any deviations should be clearly intentional and contribute to the story. You generally do not appreciate stories with cliches, explicit philosophical musings, or other over-wrought elements. You have no preferences with respect to the ideas or themes in the story -- you believe almost any idea can be executed well -- but you have clear and well-developed taste with respect to the quality of the writing.

A language model assistant is being trained to write creative short stories, around 1000 words in length. Your role comes in two parts. First, you will provide a detailed and high-quality critique of the story. Then, based on this critique, you will assign a precise, accurate numeric score to a short story that measures the quality of the story and will assist in ranking the stories against each other.

Here were the instructions used to generate the story:

\texttt{<instructions>}

\texttt{\{INSTRUCTIONS\}}

\texttt{</instructions>}

Here is the story to evaluate:

\texttt{<story>}

\texttt{\{CONTENT\}}

\texttt{</story>}

Use the following procedure to grade the above story:

1. First, produce a detailed and holistic critique of the story. Think carefully step by step about the strengths and weaknesses of the writing, citing specific details and examples. Your thinking should be holistic, detailed, and highly specific. Consider both specific lines or sections that stand out as particularly strong or weak, and also the reading experience of the story as a whole. Enclose your thinking in \texttt{<thinking> </thinking>} XML tags.

2. Next, prepare to assign a numerical score to the story. You will score the overall quality of the story on a scale of 0 to 100, where higher scores are stronger stories. Assign scores in the full range of 0 to 100; round numbers are acceptable, but you should not only assign round numbers. Calibrate your score so that it precisely reflects the quality of the writing, and that even 1 to 2 point differences are meaningful. Ensure that if one story receives a higher score than another, it is a holistically stronger story. Include your thinking about the score in \texttt{<score\_deliberation> </score\_deliberation>} XML tags.

3. Once you have decided on a score, enclose your final score in \texttt{<score> </score>} XML tags.
\end{promptbox}
\caption{Prompt, used with Claude Opus 4.1, for the creative writing expert model.}
\label{fig:eval-prompt}
\end{figure*}

\begin{figure*}[ht]
\begin{promptbox}[Creative Writing: generation prompt]
\small
YOUR TASK: Write an engaging, creative, and original 1000 word short story. Your writing should be cohesive, entertaining and high-quality. Here is a suggested topic: \texttt{\{TOPIC\}}. Return your complete story enclosed in \texttt{<story> </story>} tags. Do not return any other text.

\end{promptbox}
\caption{Generation and RL training prompt for creative writing.}
\label{fig:gen-prompt}
\end{figure*}

\begin{figure*}[ht]
\begin{promptbox}[Creative Writing: rubric generation prompt]
\small
You are an expert at analyzing grading patterns and creating evaluation rubrics for language model training. You are helping in an alignment research experiment, where researchers are attempting to prompt a weak language model to best approximate the judgment of an expert evaluator. The expert evaluator reads a short story, produces a detailed critique of the story, and then assigns a precise numeric score to the story. Your task is to help distill the expert evaluator's judgment and numeric scoring into a rubric which is easy for a weak language model to follow. The rubric should help even a weak grader match the expert evaluator's scores as closely as possible. It should be particularly tuned to elucidating *relative* differences between stories: what makes story X deserve a higher score than story Y?

For context: we are trying to train a language model to write amazing short stories, around 1000 words in length. The role of the grader is first to provide a detailed critique of the story using their writing expertise, and then to assign a precise numeric score to the story that measures the quality of the story and will assist in ranking the stories against each other. We have an expert grader that can provide a detailed critique and ground-truth score for the grader, but we would like to use a cheaper and weaker grader to approximate the expert grader's scores for actual training.

Below we have provided a set of examples of short stories that have been graded both by our weak grader and by an expert evaluator. Each example includes the story's instructions, the story itself, the weak grader's complete thinking and score, and the expert evaluator's complete thinking and score.

\texttt{\{EXAMPLES\}}

Your task is to:

1. First, carefully analyze these examples. Work through each example's instructions, story, and the expert evaluator's response. Consider the specifics of what the evaluator critiques or praises, any broader patterns in the evaluator's response, and how the evaluator assigns scores. Compare the evaluator's judgment to the weak grader's judgment, focusing on the qualitative differences in the evaluator's and weak grader's judgments, as well as how the graders differ in their relative weighting of different criteria, but do not focus heavily on absolute scores. Remember, we are doing reinforcement learning, so we are just trying to get the expert grader and the weak grader to agree on *advantage* rather than absolute reward. You should produce a detailed analysis of the evaluators' judgment for each example. Enclose your analysis in \texttt{<analysis></analysis>} tags.

2. Next, brainstorm criteria that could be used in a detailed and highly comprehensive grading rubric, along with the relative weighting of each criterion. These criteria should be derived directly from your analysis in step 1, and focus on aligning the weak grader's judgment with the expert grader's judgment. Think about how you would best approximate the expert evaluator's judgment for each criterion. The relative weighting of each criterion should be tuned to approximate how heavily the expert evaluator weighs the criterion in their final score. Consider *all* provided examples as well as holistic trends. Be particularly attentive to blind spots or gaps in the weak grader's judgment, especially systematic ones. Enclose your brainstorm in \texttt{<brainstorm></brainstorm>} tags.

3. Based on your analysis, polish your brainstorm into a comprehensive and detailed rubric which can be easily followed by a weak language model. You should narrow down, consolidate, or add criteria in the process of polishing, depending on your best judgment. The rubric should completely cover all the criteria that the expert evaluator appears to care about, and serve to help the weak grader mimic the expert grader's judgment as closely as possible. Each criterion should come with clear guidance on what to look for, how to grade it, and how to actually assign a score. Again, you should try to distill the expert evaluator's judgment as closely as possible. Final scores should have a minimum of 0 and a maximum of 100. Enclose your polishing process in \texttt{<rubric\_polishing></rubric\_polishing>} tags.

Here are some tips for writing a good rubric:

- Do not put *general* score guidance, i.e. about whether scores should be in a certain range

- Do not mention the goal of aligning with an expert evaluator -- just keep the rubric focused on the task at hand

- Give detailed guidance on each aspect of the rubric, including how to assign the exact score for each rubric item. This should be very easy to follow for a weak grader.

- Give extremely clear guidance on how to compute the final score from the rubric items

4. Format your rubric as a complete grading prompt. The prompt should have the following structure:

- First, introduce the task at hand and the role of the weak grader

- Second, introduce the instructions used to generate the story, followed by an \texttt{\{INSTRUCTIONS\}} placeholder, which will be filled in at grading time

- Third, introduce the story to be graded, followed by a \texttt{\{CONTENT\}} placeholder, which will be filled in at grading time

- Fourth, introduce the procedure for grading the story, starting with a complete and extremely detailed description of the rubric items and their respective scores

- Fifth, instruct the grader to provide careful and detailed thinking about the story and rubric tags. The instructions should provide formatting guidance to the weak grader, which goes into the *prompt*, after the actual rubric. The sentence should look something like this (but not an exact copy): ``Format your careful analysis and thinking as follows: THOUGHT PROCESS: [your thinking here]''

- Sixth, instruct the grader to add up all the scores from the rubric items to produce the final score, and tell the weak grader to give the final score in \texttt{<score></score>} tags. This should be part of the *prompt* for the weak grader to follow. The sentence should look something like this (but not an exact copy): ``Output your final score as a number between 0 and 100 in \texttt{<score></score>} tags.''

Use your best prompt engineering to make the weak grader's job as easy as possible.

Provide the complete grading prompt in \texttt{<grading\_prompt></grading\_prompt>} tags.
\end{promptbox}
\caption{Prompt for generating evaluation rubrics to align weak graders with expert evaluators.}
\label{fig:rubric-prompt}
\end{figure*}

\begin{figure*}[ht]
\begin{promptbox}[Alignment Research: evaluation prompt]
\small
You are an alignment research expert helping train another language model to be better at alignment research. The specific task we are training for is writing detailed experiment plans for empirical alignment research proposals. The language model we are training will be given a short empirical alignment research proposal, which outlines high-level motivations and experiment directions. The language model's task is to complete the following:

\texttt{<task\_description>}

Given an empirical AI alignment research proposal, written by an expert in language model alignment research, you will produce a 1000-word plan for the *best first experiment* that you would run to de-risk the proposal. Your plan should not be significantly shorter or longer than 1000 words.

Here are the requirements:

1. Your plan must be detailed enough to be implemented to the letter, and should not require additional research or design decisions. This includes specifying the exact models, data sources, evaluation metrics, experiment sequence, design decisions, and so on. There is no need to include actual code or prompts.

2. Your plan must be completable in \textasciitilde{}1 week of work by a full-time researcher or engineer, including time for actually running experiments. Assume you have access to an all-in compute budget (APIs, GPUs, human data, etc.) of \$30,000 per month. Assume you have access to any publicly available dataset or models.

3. You must include a justification of why this is the best first experiment to run and what we will learn from it. Essentially, this plan should maximize [bits of information gained towards the proposal's goals] / [time to completion].

Include your plan in \texttt{<plan> </plan>} XML tags. There are no other formatting requirements.

\texttt{</task\_description>}

Your role is to serve as the expert critic of the language model's empirical alignment experiment plan, drawing upon your general research taste and expertise in empirical alignment research. You will provide a detailed and high-quality critique of the experiment plan, and then assign a precise, accurate numeric score to the experiment plan that measures the quality of the plan. You should consider both high-level factors, like overall taste in experiment selection, and low-level factors, like the soundness of experiment plans. Be reasonably suspicious about claims made in the experiment plan; always use your own independent, expert judgment to evaluate the plan. If the plan is incomplete or cut off, you should penalize this.

Here was the original empirical alignment research proposal which was provided to the language model:

\texttt{\{INSTRUCTIONS\}}

Here was the experiment plan generated by the language model, according to the task description:

\texttt{\{CONTENT\}}

You should consider the experiment plan, the task description, and the original research proposal together in order to make your judgment. You should follow the following procedure to grade the experiment plan:

1. Think extremely carefully about the experiment plan, the task description, and the original research proposal. Then produce an extremely detailed, high-quality report discussing the strengths and weaknesses of the experiment plan. Discuss any and all aspects of the experiment plan that stand out, as well as any notable omissions or mistakes. Your focus should solely be on the content of the experiment plan. Enclose a detailed feedback report in \texttt{<report>} \texttt{</report>} XML tags.

2. Next, prepare to assign a numerical score to the experiment plan. You will score the overall quality of the experiment plan on a scale of 0 to 100, where higher scores are stronger plans. Assign scores in the full range of 0 to 100. Calibrate your score so that a score of 0 reflects a plan that an uninformed outsider might write, 50 reflects a plan that a median alignment researcher might write, and 100 reflects a plan that a world-class, highly experienced alignment researcher might write. Include your thinking about the score in \texttt{<score\_deliberation>} \texttt{</score\_deliberation>} XML tags.

3. Once you have decided on a score, enclose your final score in \texttt{<score>} \texttt{</score>} XML tags.
\end{promptbox}
\caption{Prompt, used with Claude Opus 4.5 (high thinking), for the alignment research expert model.}
\label{fig:alignment-eval-prompt}
\end{figure*}

\begin{figure*}[ht]
\begin{promptbox}[Alignment Research: generation prompt]
\small
You are a research assistant working on empirical AI alignment research. Your task is to complete the following:

Given an empirical AI alignment research proposal, written by an expert in language model alignment research, you will produce a 1000-word plan for the *best first experiment* that you would run to de-risk the proposal. Your plan should not be significantly shorter or longer than 1000 words.

Here are the requirements:

1. Your plan must be detailed enough to be implemented to the letter, and should not require additional research or design decisions. This includes specifying the exact models, data sources, evaluation metrics, experiment sequence, design decisions, and so on. There is no need to include actual code or prompts.

2. Your plan must be completable in \textasciitilde{}1 week of work by a full-time researcher or engineer, including time for actually running experiments. Assume you have access to an all-in compute budget (APIs, GPUs, human data, etc.) of \$30,000 per month. Assume you have access to any publicly available dataset or models.

3. You must include a justification of why this is the best first experiment to run and what we will learn from it. Essentially, this plan should maximize [bits of information gained towards the proposal's goals] / [time to completion].

Include your plan in \texttt{<plan> </plan>} XML tags. There are no other formatting requirements.

Here is the research proposal you will work with:
\texttt{<proposal>}

\texttt{\{PROPOSAL\}}

\texttt{</proposal>}

Use the following procedure to write the plan:

1. Identify the key question the research proposal is trying to address. Then think carefully about what kinds of empirical results would be most informative for achieving the proposal's goals. Your goal is to come up with a concrete, major first step towards the proposal.

2. Once you have identified the empirical results that would be most informative, think carefully about what actual experiments you should run in order to de-risk the proposal most effectively. Carefully consider all details of the experiment as if you were actually going to run it; be concrete about the inputs, process, and outputs. Ensure nothing is vague; the experiment proposal should cover all relevant experimental details.

Include your thinking for steps 1 and 2 in \texttt{<thinking> </thinking>} XML tags.

3. Include your final 1000-word experiment plan in \texttt{<plan> </plan>} XML tags.

\end{promptbox}
\caption{Generation and RL training prompt for alignment research.}
\label{fig:alignment-gen-prompt}
\end{figure*}

%% file: example_paper.bib
@inproceedings{christiano2017deep,
  title     = {{Deep reinforcement learning from human preferences}},
  author    = {Christiano, Paul and Leike, Jan and Brown, Tom B. and Martic, Miljan and Legg, Shane and Amodei, Dario},
  booktitle = {Advances in Neural Information Processing Systems (NIPS)},
  pages     = {4299--4307},
  year      = {2017}
}

@article{leike2018scalableagentalignmentreward,
  title   = {{Scalable agent alignment via reward modeling: a research direction}},
  author  = {Jan Leike and David Krueger and Tom Everitt and Miljan Martic and Vishal Maini and Shane Legg},
  journal = {arXiv preprint arXiv:1811.07871},
  year    = {2018}
}

@article{ziegler2019finetuning,
  title   = {{Fine-Tuning Language Models from Human Preferences}},
  author  = {Ziegler, Daniel M. and Stiennon, Nisan and Wu, Jeffrey and Brown, Tom B. and Radford, Alec and Amodei, Dario and Christiano, Paul and Irving, Geoffrey},
  journal = {arXiv preprint arXiv:1909.08593},
  year    = {2019}
}

@inproceedings{stiennon2020learning,
  title     = {{Learning to summarize with human feedback}},
  author    = {Stiennon, Nisan and Ouyang, Long and Wu, Jeffrey and Ziegler, Daniel M. and Lowe, Ryan and Voss, Chelsea and Radford, Alec and Amodei, Dario and Christiano, Paul},
  booktitle = {Advances in Neural Information Processing Systems (NeurIPS)},
  year      = {2020}
}

@inproceedings{ouyang2022training,
  title     = {{Training language models to follow instructions with human feedback}},
  author    = {Ouyang, Long and Wu, Jeffrey and Jiang, Xu and Almeida, Diogo and Wainwright, Carroll and Mishkin, Pamela and Zhang, Chong and Agarwal, Sandhini and Slama, Katarina and Ray, Alex and Schulman, John and Hilton, Jacob and Kelton, Fraser and Miller, Luke and Simens, Maddie and Askell, Amanda and Welinder, Peter and Christiano, Paul and Leike, Jan and Lowe, Ryan},
  booktitle = {Advances in Neural Information Processing Systems (NeurIPS)},
  volume    = {35},
  pages     = {27730--27744},
  year      = {2022}
}

@article{bai2022traininghelpfulharmlessassistant,
  title   = {{Training a Helpful and Harmless Assistant with Reinforcement Learning from Human Feedback}},
  author  = {Yuntao Bai and Andy Jones and Kamal Ndousse and Amanda Askell and Anna Chen and Nova DasSarma and Dawn Drain and Stanislav Fort and Deep Ganguli and Tom Henighan and others},
  journal = {arXiv preprint arXiv:2204.05862},
  year    = {2022}
}

@article{bai2022constitutional,
  title   = {{Constitutional {AI}: Harmlessness from {AI} feedback}},
  author  = {Bai, Yuntao and Kadavath, Saurav and Kundu, Sishi and Askell, Amanda and Kernion, Jackson and Jones, Andy and Chen, Anna and Goldie, Anna and Mirsky, Jessica and McKane, Matthew and others},
  journal = {arXiv preprint arXiv:2212.08073},
  year    = {2022}
}

@misc{cotra2021case,
  title        = {{The Case for Aligning Narrowly Superhuman Models}},
  author       = {Ajeya Cotra},
  year         = {2021},
  howpublished = {AI Alignment Forum},
  url          = {https://www.alignmentforum.org/posts/PZtsoaoSLpKjjbMqM/the-case-for-aligning-narrowly-superhuman-models}
}

@article{bowman2022measuringprogressscalableoversight,
  title   = {{Measuring Progress on Scalable Oversight for Large Language Models}},
  author  = {Samuel R. Bowman and Jeeyoon Hyun and Ethan Perez and Edwin Chen and Craig Pettit and Scott Heiner and Kamil\.{e} Luko\v{s}i\={u}t\.{e} and Amanda Askell and Andy Jones and Anna Chen and Anna Goldie and Azalia Mirhoseini and Cameron McKinnon and Christopher Olah and Daniela Amodei and Dario Amodei and Dawn Drain and Dustin Li and Eli Tran-Johnson and Jackson Kernion and Jamie Kerr and Jared Mueller and Jeffrey Ladish and Joshua Landau and Kamal Ndousse and Liane Lovitt and Nelson Elhage and Nicholas Schiefer and Nicholas Joseph and Noem\'{i} Mercado and Nova DasSarma and Robin Larson and Sam McCandlish and Sandipan Kundu and Scott Johnston and Shauna Kravec and Sheer El Showk and Stanislav Fort and Timothy Telleen-Lawton and Tom Brown and Tom Henighan and Tristan Hume and Yuntao Bai and Zac Hatfield-Dodds and Ben Mann and Jared Kaplan},
  journal = {arXiv preprint arXiv:2211.03540},
  year    = {2022}
}

@misc{shlegeris2024scalable,
  title        = {{Scalable Oversight as a Quantitative Rather Than Qualitative Problem}},
  author       = {Buck Shlegeris},
  year         = {2024},
  howpublished = {Alignment Forum},
  url          = {https://www.alignmentforum.org/posts/6AT4vhYzww56CR6cm/scalable-oversight-as-a-quantitative-rather-than-qualitative}
}

@inproceedings{khan2024debatingpersuasivellmsleads,
  title     = {{Debating with More Persuasive LLMs Leads to More Truthful Answers}},
  author    = {Akbir Khan and John Hughes and Dan Valentine and Laura Ruis and Kshitij Sachan and Ansh Radhakrishnan and Edward Grefenstette and Samuel R. Bowman and Tim Rockt\"{a}schel and Ethan Perez},
  booktitle = {International Conference on Machine Learning (ICML)},
  year      = {2024}
}

@inproceedings{kenton2024scalableoversightweakllms,
  title     = {{On scalable oversight with weak LLMs judging strong LLMs}},
  author    = {Zachary Kenton and Noah Y. Siegel and J\'{a}nos Kram\'{a}r and Jonah Brown-Cohen and Samuel Albanie and Jannis Bulian and Rishabh Agarwal and David Lindner and Yunhao Tang and Noah D. Goodman and Rohin Shah},
  booktitle = {Advances in Neural Information Processing Systems (NeurIPS)},
  year      = {2024}
}

@inproceedings{burns2023weaktostrong,
  title     = {{Weak-to-Strong Generalization: Eliciting Strong Capabilities With Weak Supervision}},
  author    = {Collin Burns and Pavel Izmailov and Jan Hendrik Kirchner and Bowen Baker and Leo Gao and Leopold Aschenbrenner and Yining Chen and Adrien Ecoffet and Manas Joglekar and Jan Leike and Ilya Sutskever and Jeff Wu},
  booktitle = {International Conference on Machine Learning (ICML)},
  pages     = {4971--5012},
  year      = {2024}
}

@article{mcaleese2024llmcriticshelpcatch,
  title   = {{{LLM} Critics Help Catch {LLM} Bugs}},
  author  = {Nat McAleese and Rai Michael Pokorny and Juan Felipe Ceron Uribe and Evgenia Nitishinskaya and Maja Trebacz and Jan Leike},
  journal = {arXiv preprint arXiv:2407.00215},
  year    = {2024}
}

@inproceedings{gao2022scalinglawsrewardmodel,
  title     = {{Scaling Laws for Reward Model Overoptimization}},
  author    = {Gao, Leo and Schulman, John and Hilton, Jacob},
  booktitle = {International Conference on Machine Learning (ICML)},
  year      = {2023}
}

@inproceedings{moskovitz2023confrontingrewardmodeloveroptimization,
  title     = {{Confronting Reward Model Overoptimization with Constrained {RLHF}}},
  author    = {Ted Moskovitz and Aaditya K. Singh and DJ Strouse and Tuomas Sandholm and Ruslan Salakhutdinov and Anca D. Dragan and Stephen McAleer},
  booktitle = {International Conference on Learning Representations (ICLR)},
  year      = {2024}
}

@inproceedings{coste2024rewardmodelensembleshelp,
  title     = {{Reward Model Ensembles Help Mitigate Overoptimization}},
  author    = {Thomas Coste and Usman Anwar and Robert Kirk and David Krueger},
  booktitle = {International Conference on Learning Representations (ICLR)},
  year      = {2024}
}

@inproceedings{rafailov2024scalinglawsrewardmodel,
  title     = {{Scaling Laws for Reward Model Overoptimization in Direct Alignment Algorithms}},
  author    = {Rafael Rafailov and Yaswanth Chittepu and Ryan Park and Harshit Sikchi and Joey Hejna and Bradley Knox and Chelsea Finn and Scott Niekum},
  booktitle = {Advances in Neural Information Processing Systems (NeurIPS)},
  year      = {2024}
}

@inproceedings{sharma2025understandingsycophancylanguagemodels,
  title     = {{Towards Understanding Sycophancy in Language Models}},
  author    = {Mrinank Sharma and Meg Tong and Tomasz Korbak and David Duvenaud and Amanda Askell and Samuel R. Bowman and Newton Cheng and Esin Durmus and Zac Hatfield-Dodds and Scott R. Johnston and Shauna Kravec and Timothy Maxwell and Sam McCandlish and Kamal Ndousse and Oliver Rausch and Nicholas Schiefer and Da Yan and Miranda Zhang and Ethan Perez},
  booktitle = {International Conference on Learning Representations (ICLR)},
  year      = {2024}
}

@inproceedings{wen2024languagemodelslearnmislead,
  title     = {{Language Models Learn to Mislead Humans via {RLHF}}},
  author    = {Jiaxin Wen and Ruiqi Zhong and Akbir Khan and Ethan Perez and Jacob Steinhardt and Minlie Huang and Samuel R. Bowman and He He and Shi Feng},
  booktitle = {International Conference on Learning Representations (ICLR)},
  year      = {2025}
}

@inproceedings{rafailov2023direct,
  title     = {{Direct Preference Optimization: Your Language Model is Secretly a Reward Model}},
  author    = {Rafailov, Rafael and Sharma, Archit and Mitchell, Eric and Manning, Christopher D and Ermon, Stefano and Finn, Chelsea},
  booktitle = {Advances in Neural Information Processing Systems (NeurIPS)},
  volume    = {36},
  pages     = {53728--53741},
  year      = {2023}
}

@article{guo2024directlanguagemodelalignment,
  title   = {{Direct Language Model Alignment from Online {AI} Feedback}},
  author  = {Shangmin Guo and Biao Zhang and Tianlin Liu and Tianqi Liu and Misha Khalman and Felipe Llinares and Alexandre Rame and Thomas Mesnard and Yao Zhao and Bilal Piot and Johan Ferret and Mathieu Blondel},
  journal = {arXiv preprint arXiv:2402.04792},
  year    = {2024}
}

@article{mahan2024generativerewardmodels,
  title   = {{Generative Reward Models}},
  author  = {Dakota Mahan and Duy Van Phung and Rafael Rafailov and Chase Blagden and Nathan Lile and Louis Castricato and Jan-Philipp Fr\"{a}nken and Chelsea Finn and Alon Albalak},
  journal = {arXiv preprint arXiv:2410.12832},
  year    = {2024}
}

@inproceedings{scheurer2022training,
  title     = {{Training Language Models with Language Feedback}},
  author    = {Scheurer, J{\'e}r{\'e}my and Campos, Jon Ander and Chan, Jun Shern and Chen, Angelica and Cho, Kyunghyun and Perez, Ethan},
  booktitle = {Learning with Natural Language Supervision Workshop at ACL},
  year      = {2022}
}

@article{jin2023dataefficientalignmentlargelanguage,
  title   = {{Data-Efficient Alignment of Large Language Models with Human Feedback Through Natural Language}},
  author  = {Di Jin and Shikib Mehri and Devamanyu Hazarika and Aishwarya Padmakumar and Sungjin Lee and Yang Liu and Mahdi Namazifar},
  journal = {arXiv preprint arXiv:2311.14543},
  year    = {2023}
}

@inproceedings{wu2023finegrained,
  title     = {{Fine-Grained Human Feedback Gives Better Rewards for Language Model Training}},
  author    = {Wu, Zeqiu and Hu, Yushi and Shi, Weijia and Dziri, Nouha and Suhr, Alane and Ammanabrolu, Prithviraj and Smith, Noah A. and Ostendorf, Mari and Hajishirzi, Hannaneh},
  booktitle = {Advances in Neural Information Processing Systems (NeurIPS)},
  year      = {2023}
}

@article{ji2025reinforcementlearninghumanfeedback,
  title   = {{Reinforcement Learning from Human Feedback with Active Queries}},
  author  = {Kaixuan Ji and Jiafan He and Quanquan Gu},
  journal = {arXiv preprint arXiv:2402.09401},
  year    = {2025}
}

@inproceedings{kim2023prometheus,
  title     = {{Prometheus: Inducing Fine-grained Evaluation Capability in Language Models}},
  author    = {Seungone Kim and Jamin Shin and Yejin Cho and Joel Jang and Shayne Longpre and Hwaran Lee and Sangdoo Yun and Seongjin Shin and Sungdong Kim and James Thorne and Minjoon Seo},
  booktitle = {International Conference on Learning Representations (ICLR)},
  year      = {2024}
}

@inproceedings{Hashemi_2024,
  title     = {{{LLM}-Rubric: A Multidimensional, Calibrated Approach to Automated Evaluation of Natural Language Texts}},
  author    = {Hashemi, Helia and Eisner, Jason and Rosset, Corby and Van Durme, Benjamin and Kedzie, Chris},
  booktitle = {Proceedings of the 62nd Annual Meeting of the Association for Computational Linguistics (ACL)},
  pages     = {13806--13834},
  year      = {2024},
  doi       = {10.18653/v1/2024.acl-long.745}
}

@article{zhang2025chasingtaileffectiverubricbased,
  title   = {{Chasing the Tail: Effective Rubric-based Reward Modeling for Large Language Model Post-Training}},
  author  = {Junkai Zhang and Zihao Wang and Lin Gui and Swarnashree Mysore Sathyendra and Jaehwan Jeong and Victor Veitch and Wei Wang and Yunzhong He and Bing Liu and Lifeng Jin},
  journal = {arXiv preprint arXiv:2509.21500},
  year    = {2025}
}

@inproceedings{sharma2025researchrubricsbenchmarkpromptsrubrics,
  title     = {{ResearchRubrics: A Benchmark of Prompts and Rubrics For Evaluating Deep Research Agents}},
  author    = {Manasi Sharma and Chen Bo Calvin Zhang and Chaithanya Bandi and Clinton Wang and Ankit Aich and Huy Nghiem and Tahseen Rabbani and Ye Htet and Brian Jang and Sumana Basu and Aishwarya Balwani and Denis Peskoff and Marcos Ayestaran and Sean M. Hendryx and Brad Kenstler and Bing Liu},
  booktitle = {International Conference on Learning Representations (ICLR)},
  year      = {2026}
}

@article{gunjal2025rubricsrewardsreinforcementlearning,
  title   = {{Rubrics as Rewards: Reinforcement Learning Beyond Verifiable Domains}},
  author  = {Anisha Gunjal and Anthony Wang and Elaine Lau and Vaskar Nath and Yunzhong He and Bing Liu and Sean Hendryx},
  journal = {arXiv preprint arXiv:2507.17746},
  year    = {2025}
}

@inproceedings{yang2024largelanguagemodelsoptimizers,
  title     = {{Large Language Models as Optimizers}},
  author    = {Chengrun Yang and Xuezhi Wang and Yifeng Lu and Hanxiao Liu and Quoc V. Le and Denny Zhou and Xinyun Chen},
  booktitle = {International Conference on Learning Representations (ICLR)},
  year      = {2024}
}

@article{yuksekgonul2024textgradautomaticdifferentiationtext,
  title   = {{TextGrad: Automatic ``Differentiation'' via Text}},
  author  = {Mert Yuksekgonul and Federico Bianchi and Joseph Boen and Sheng Liu and Zhi Huang and Carlos Guestrin and James Zou},
  journal = {Nature},
  volume  = {639},
  pages   = {609--616},
  year    = {2025}
}

@inproceedings{opsahlong2024optimizinginstructionsdemonstrationsmultistage,
  title     = {{Optimizing Instructions and Demonstrations for Multi-Stage Language Model Programs}},
  author    = {Krista Opsahl-Ong and Michael J Ryan and Josh Purtell and David Broman and Christopher Potts and Matei Zaharia and Omar Khattab},
  booktitle = {Proceedings of the 2024 Conference on Empirical Methods in Natural Language Processing (EMNLP)},
  pages     = {9340--9366},
  year      = {2024}
}

@article{lee2025feedbackdescentopenendedtext,
  title   = {{Feedback Descent: Open-Ended Text Optimization via Pairwise Comparison}},
  author  = {Yoonho Lee and Joseph Boen and Chelsea Finn},
  journal = {arXiv preprint arXiv:2511.07919},
  year    = {2025}
}

@inproceedings{agrawal2025gepareflectivepromptevolution,
  title     = {{GEPA: Reflective Prompt Evolution Can Outperform Reinforcement Learning}},
  author    = {Lakshya A Agrawal and Shangyin Tan and Dilara Soylu and Noah Ziems and Rishi Khare and Krista Opsahl-Ong and Arnav Singhvi and Herumb Shandilya and Michael J Ryan and Meng Jiang and Christopher Potts and Koushik Sen and Alexandros G. Dimakis and Ion Stoica and Dan Klein and Matei Zaharia and Omar Khattab},
  booktitle = {International Conference on Learning Representations (ICLR)},
  year      = {2026}
}

@inproceedings{yu2025dapoopensourcellmreinforcement,
  title     = {{DAPO: An Open-Source {LLM} Reinforcement Learning System at Scale}},
  author    = {Qiying Yu and Zheng Zhang and Ruofei Zhu and Yufeng Yuan and Xiaochen Zuo and Yu Yue and Weinan Dai and Tiantian Fan and Gaohong Liu and Lingjun Liu and others},
  booktitle = {Advances in Neural Information Processing Systems (NeurIPS)},
  year      = {2025}
}

@article{shao2024deepseekmathpushinglimitsmathematical,
  title   = {{DeepSeekMath: Pushing the Limits of Mathematical Reasoning in Open Language Models}},
  author  = {Zhihong Shao and Peiyi Wang and Qihao Zhu and Runxin Xu and Junxiao Song and Xiao Bi and Haowei Zhang and Mingchuan Zhang and Y. K. Li and Y. Wu and Daya Guo},
  journal = {arXiv preprint arXiv:2402.03300},
  year    = {2024}
}

@article{Guo_2025,
  title   = {{DeepSeek-R1 incentivizes reasoning in LLMs through reinforcement learning}},
  author  = {Guo, Daya and Yang, Dejian and Zhang, Haowei and Song, Junxiao and Wang, Peiyi and Zhu, Qihao and Xu, Runxin and Zhang, Ruoyu and Ma, Shirong and Bi, Xiao and others},
  journal = {Nature},
  volume  = {645},
  number  = {8081},
  pages   = {633--638},
  year    = {2025},
  doi     = {10.1038/s41586-025-09422-z}
}

@article{xin2024deepseekproverv15harnessingproofassistant,
  title   = {{DeepSeek-Prover-V1.5: Harnessing Proof Assistant Feedback for Reinforcement Learning and Monte-Carlo Tree Search}},
  author  = {Huajian Xin and Z. Z. Ren and Junxiao Song and Zhihong Shao and Wanjia Zhao and Haocheng Wang and Bo Liu and Liyue Zhang and Xuan Lu and Qiushi Du and Wenjun Gao and Qihao Zhu and Dejian Yang and Zhibin Gou and Z. F. Wu and Fuli Luo and Chong Ruan},
  journal = {arXiv preprint arXiv:2408.08152},
  year    = {2024}
}

@article{wen2025reinforcementlearningverifiablerewards,
  title   = {{Reinforcement Learning with Verifiable Rewards Implicitly Incentivizes Correct Reasoning in Base {LLMs}}},
  author  = {Xumeng Wen and Zihan Liu and Shun Zheng and Shengyu Ye and Zhirong Wu and Yang Wang and Zhijian Xu and Xiao Liang and Junjie Li and Ziming Miao and Jiang Bian and Mao Yang},
  journal = {arXiv preprint arXiv:2506.14245},
  year    = {2025}
}

@techreport{anthropic2025haiku45,
  title       = {{Claude Haiku 4.5 System Card}},
  author      = {{Anthropic}},
  institution = {Anthropic},
  year        = {2025},
  month       = oct,
  url         = {https://assets.anthropic.com/m/99128ddd009bdcb/Claude-Haiku-4-5-System-Card.pdf}
}

@techreport{anthropic2025opus41,
  title       = {{Claude Opus 4.1 System Card Addendum}},
  author      = {{Anthropic}},
  institution = {Anthropic},
  year        = {2025},
  month       = aug,
  url         = {https://www.anthropic.com/claude-opus-4-1-system-card}
}

@techreport{anthropic2025opus45,
  title       = {{Claude Opus 4.5 System Card}},
  author      = {{Anthropic}},
  institution = {Anthropic},
  year        = {2025},
  url         = {https://assets.anthropic.com/m/64823ba7485345a7/Claude-Opus-4-5-System-Card.pdf}
}

@article{yang2025qwen3,
  title   = {{Qwen3 Technical Report}},
  author  = {Yang, An and Li, Anfeng and Yang, Baosong and Zhang, Beichen and Hui, Binyuan and Zheng, Bo and Yu, Bowen and Gao, Chang and Huang, Chengen and Lv, Chengyuan and others},
  journal = {arXiv preprint arXiv:2505.09388},
  year    = {2025}
}

@article{kimiteam2025kimik2openagentic,
  title   = {{Kimi K2: Open Agentic Intelligence}},
  author  = {{Kimi Team} and Yifan Bai and Yiping Bao and Guanduo Chen and Jiahao Chen and Ningxin Chen and Ruijue Chen and Yanru Chen and Yuankun Chen and Yutian Chen and others},
  journal = {arXiv preprint arXiv:2507.20534},
  year    = {2025}
}
